\documentclass[3p, 12pt]{elsarticle}

\usepackage{amssymb}
\usepackage{amsmath}

\usepackage{multirow}
\usepackage{xcolor}

\usepackage{caption}
\usepackage{subcaption}
\usepackage{svg}

\PassOptionsToPackage{hyphens}{url}\usepackage{hyperref}

\hypersetup{
    colorlinks=true,
    linkcolor=blue,
    filecolor=blue,      
    urlcolor=blue,
    citecolor=blue,
}
\journal{Applied Energy}
\date{}
\makeatletter % remove preprint
\def\ps@pprintTitle{%
 \let\@oddhead\@empty
 \let\@evenhead\@empty
 \def\@oddfoot{}%
 \let\@evenfoot\@oddfoot}
\makeatother

\begin{document}

\begin{frontmatter}



\title{Day-ahead regional solar power forecasting with hierarchical temporal convolutional neural networks using historical power generation and weather data}
% \title{Regional Solar Power Forecasting Using Hierarchical Temporal Convolutional Neural Networks}

\author[unimelb]{Maneesha Perera\corref{cor1}}
\ead{maneesha.perera1@unimelb.edu.au}
\author[unimelb]{Julian De Hoog}
\ead{julian.dehoog@unimelb.edu.au}
\author[unimelb2]{Kasun Bandara}
\ead{kasun.bandara@unimelb.edu.au}
\author[unimelb]{Damith Senanayake}
\ead{damith.senanayake@unimelb.edu.au}
\author[unimelb]{Saman Halgamuge}
\ead{saman@unimelb.edu.au}

\address[unimelb]{Department of Mechanical Engineering, School of Electrical, Mechanical and Infrastructure Engineering, The University of Melbourne, Melbourne, Australia}
\address[unimelb2]{School of Computing and Information Systems, Melbourne Centre for Data Science, The University of Melbourne, Melbourne, Australia}
\cortext[cor1]{Corresponding author}

\nonumnote{This manuscript was accepted for publication on March 3, 2024. This is the accepted version of the manuscript.}
\nonumnote{© 2024. This manuscript version is made available under the CC-BY-NC-ND 4.0 license \url{https://creativecommons.org/licenses/by-nc-nd/4.0/}}

\begin{abstract}
% Abstract

Regional solar power forecasting, which involves predicting the total power generation from all rooftop photovoltaic (PV) systems in a region holds significant importance for various stakeholders in the energy sector to ensure a stable electricity supply. However, the vast amount of solar power generation and weather time series from geographically dispersed locations that need to be considered in the forecasting process makes accurate regional forecasting challenging. Therefore, previous studies have limited the focus to either forecasting a single time series (i.e., aggregated time series) which is the addition of all solar generation time series in a region, disregarding the location-specific weather effects or forecasting solar generation time series of each PV site (i.e., individual time series) independently using location-specific weather data, resulting in a large number of forecasting models. In this work, we propose two new deep-learning-based regional forecasting methods that can effectively leverage both types of time series (aggregated and individual) with weather data in a region. We propose two \underline{h}ierarchical \underline{t}emporal \underline{c}onvolutional \underline{n}eural \underline{n}etwork architectures (HTCNN A1 and A2) and two new strategies to adapt HTCNNs for regional solar power forecasting. In the first strategy, we explore generating a regional forecast using a single HTCNN. In the second, we divide the region into multiple sub-regions based on weather information and train separate HTCNNs for each sub-region; the forecasts of each sub-region are then added to generate a regional forecast. The proposed work is evaluated using a large dataset collected over a year from 101 locations across Western Australia to provide a day ahead forecast at an hourly time resolution which involves forecasting a horizon of 18 hours. We compare our approaches with well-known alternative methods, including long-short term memory networks and convolution neural networks, and show that proposed HTCNN-based approaches require fewer individually trained networks. Furthermore, the sub-region-based HTCNN-based approach achieves a forecast skill score of 40.2\% and reduces a statistically significant forecast error by 6.5\% compared to the best-performing counterpart. Our results indicate that the proposed approaches are well-suited for forecasting applications covering large regions containing many individual solar PV systems in different locations.

\end{abstract}
\end{frontmatter}




% \begin{highlights}
% \end{highlights}



\section{Introduction}
\label{sec:intro}

In recent years, distributed solar photovoltaic (PV) systems on the rooftops of homes have rapidly increased \cite{iea, csiro}. As a result, many energy sector participants need to manage multiple emerging issues from such high levels of rooftop solar installations \cite{aemo, Koster2019Short-termLuxembourg, Bright2018ImprovedSystems}. For instance, transmission system operators who manage electricity supply and demand must account for voltage fluctuations, oversupply of electricity, and sudden ramping needs. Furthermore, market operators who govern and manage energy markets must accurately forecast the net demand within the regions they govern, considering the impact of distributed solar on the net demand. Consequently, accurate forecasts of the power generation from distributed solar PV systems are vital to alleviate many of these issues faced by energy participants \cite{Yang2019OperationalMarket, Yang2021OperationalOutlook, BrancucciMartinez-Anido2016TheImprovement}.

There has been a significant effort in the solar forecasting community to forecast the power output of individual PV systems and utility-scale solar plants, either by directly forecasting the power output using historical power generation data, or by forecasting solar irradiance and converting it to electrical power \cite{Antonanzas2016ReviewForecasting}. However, for both grid and market operators, it is becoming increasingly important to have regional solar forecasts that consider the power generation of the full cohort of distributed PV systems across a whole region \cite{Koster2019Short-termLuxembourg, Antonanzas2016ReviewForecasting, Huertas-Tato2020ALearning}. The total solar power output of a region can depend on the solar power output of hundreds, or even thousands, of individual PV systems installed in geographically dispersed locations, experiencing differing weather conditions \cite{DaSilvaFonsecaJunior2014RegionalAnalysis}. While nearby systems may have similar generation profiles, geographically far-apart systems can have significantly different generation patterns. These different power generation profiles and weather conditions must all be considered to achieve an accurate regional forecast. Furthermore, in many parts of the world, time series data of individual PV systems are becoming increasingly available due to the increased installation of metering infrastructure \cite{Koster2019Short-termLuxembourg}. Therefore, it is becoming important to understand how such a large volume of time series datasets can be used to improve the regional distributed solar generation forecasts. 

\begin{figure*}[]
    \centering
    \includegraphics[width=0.95\textwidth]{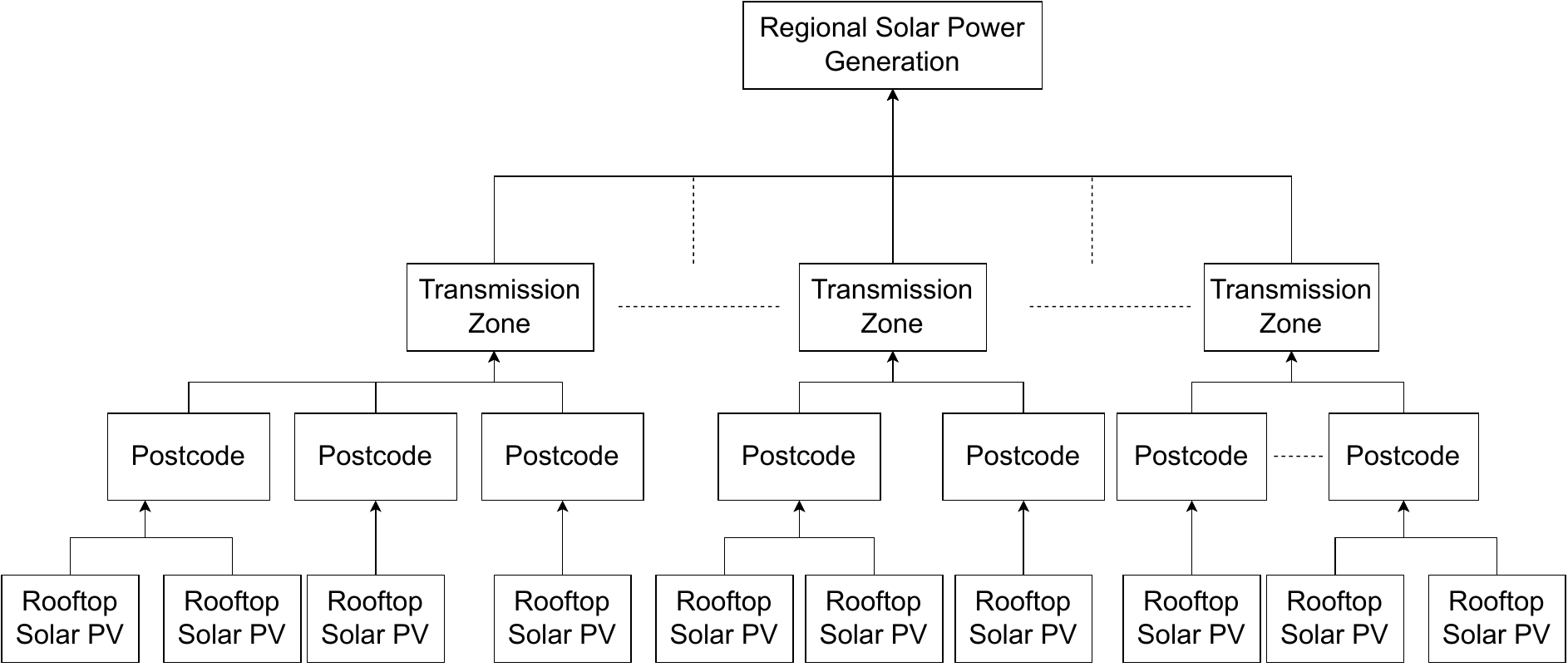}
    \caption{The hierarchical structure of solar power generation time series datasets in a region.}
    \label{fig:heirarchy}
\end{figure*}

Figure \ref{fig:heirarchy} shows a hierarchical structure that can be used to represent such time series datasets available in a region. Power generation data from nearby rooftop PV systems can be grouped into multiple small areas such as by postcode or transmission zone level. The regional solar power generation reflects the total solar power generation of the region. In such a hierarchy time series at each level is an addition of its associated bottom level series. Existing regional solar forecasting work using power generation time series datasets often make use of the regional solar power generation time series (i.e., top level time series of the hierarchy) \cite{Almaghrabi2021ForecastingApproach, Almaghrabi2021SpatiallyNetworks, Rana2020AProduction, Yang2017ReconcilingHierarchy} or the power generation time series from all rooftop PV systems (i.e., bottom level time series of the hierarchy) \cite{Koster2019Short-termLuxembourg, DaSilvaFonseca2014RegionalMethods, Zhang2020AForecast, Yu2020ImprovedForecast}. However, as discussed in prior work \cite{Hollyman2021UnderstandingReconciliation}, each time series in such a hierarchy can have its own unique features that can be leveraged to improve the forecasts in other levels. For instance, time series at bottom levels may include patterns that are more closely related to local weather data than regional power generation. Furthermore, the regional power generation time series may have smooth patterns due to the averaging effect of hundreds or thousands of individual power generation time series \cite{Antonanzas2016ReviewForecasting, Huertas-Tato2020ALearning}.

Among various forecasting methods, deep learning methods have gained particular interest in forecasting solar generation due to the ability of these methods to capture complex non-linear relationships between solar power generation and associated weather data \cite{Wang2019ANetwork}. Convolutional neural networks (CNNs) are a class of deep neural networks (DNN) that can automatically learn and extract relevant features for a specific task without manual feature engineering or prior knowledge of the task \cite{Huang2017DenselyNetworks, he2016deep}. While these networks were originally designed for computer vision applications, in recent years they have shown promising results when applied to solar power forecasting \cite{Wang2019ANetwork, Heo2021Multi-channelForecasting}. 

In this work, we propose two DNN architectures for regional solar power forecasting, referred to herein as \underline{H}ierarchical \underline{T}emporal \underline{C}onvolutional \underline{N}eural \underline{N}etworks (HTCNNs), that can model aggregated and individual power generation time series with weather data in a single neural network. Throughout this work, we consider the power generation datasets from an intermediate level of the time series hierarchy, such as postcode level (referred to herein as the individual power generation data), instead of the rooftop solar level, as there can be a high level of noise associated with a finer-grain level \cite{Hollyman2021UnderstandingReconciliation} that may not be beneficial in forecasting the regional solar generation which is the top most level in the hierarchy. HTCNNs are based on a type of CNNs known as temporal convolutional neural networks (TCNs). Although long short-term memory networks (LSTMs) are frequently used for time-series forecasting, TCNs are faster to train and can capture longer-term temporal features compared to LSTMs and therefore are chosen as the basic building block in this work \cite{Lea2017TemporalDetection, Oord2016WaveNet:Audio, Bai2018AnModeling, Lin2020TemporalForecasting}. HTCNNs are hierarchical with the implication that they are designed for time series with a hierarchical structure. 

% To the best of our knowledge, this is the first study that attempts to forecast an aggregated solar power generation time series by modelling the aggregated, individual power generation and weather time series coming from geographically dispersed locations using a single network.

In order to understand how HTCNNs can be applied to the solar power forecasting task, we propose two strategies. The first strategy directly forecasts the power generation of the entire region using a single HTCNN. The second strategy divides the region into sub-regions based on the available local weather information. Each sub-region consists of multiple individual power generation time series affected by similar weather conditions as they are geographically located close to each other. The aggregated power generation of each sub-region is forecasted separately using a HTCNN. Finally, these forecasts are added to achieve the forecast for the entire region of interest. We evaluate the proposed work in comparison to well-known statistical and deep learning baselines for forecasting, including periodic persistence, autoregressive models, CNNs, TCNs, and LSTMs. Each benchmark forecasting model is adapted for regional solar power forecasting using two existing regional forecasting strategies where either the aggregated power generation data or the individual time series data (postcode-level data) with associated weather data are used to make a regional forecast.

In summary, the main contributions of this work are as follows:
\begin{enumerate}
    \item We propose two DNNs \textit{HTCNN A1}, and \textit{A2} to forecast an aggregated power generation time series by jointly modelling the aggregated power generation time series, the individual power generation time series, and multiple weather time series in a single neural network.
     \item We introduce two strategies to use HTCNNs for regional solar power forecasting.
    \item We compare the performance of the proposed HTCNNs-based approaches to baseline methods using a solar PV power dataset collected from 101 diverse locations in the Southwest Interconnected System (SWIS) in Western Australia. We show that our approaches can significantly reduce the regional forecast error while reducing the number of networks needed to achieve an accurate forecast. Furthermore, our results provide insights into the ability of DNNs to learn across multiple time series datasets available in the region.
\end{enumerate}

The rest of the paper is structured as follows. In Section \ref{sec:related}, related work is discussed, in Section \ref{sec:method} the (i) advantages of jointly modelling aggregated and individual time series datasets, (ii) main components of TCNs, (iii) components of the HTCNNs and (iv) regional forecasting strategies with HTCNNs are presented. In Section \ref{sec:case_study}, data, baseline forecasting models and the evaluation is outlined. In Section \ref{sec:results}, the results are presented with a discussion. Finally, in Section \ref{sec:conclusion}, the conclusions of this work are presented.

\section{Related Work}
\label{sec:related}
Studies related to solar irradiance and solar PV power forecasting have seen rapid progress in recent years \cite{Yang2018HistoryMining, Hong2020EnergyOutlook}. Many forecasting methods exist for solar forecasting, and reviews of these techniques can be found in: \cite{Antonanzas2016ReviewForecasting, Yang2018HistoryMining, Raza2016OnForecast, Ahmed2020AOptimization}. In the following, we discuss related work in the area of (i) deep learning methods used in solar forecasting and (ii,) approaches for regional solar power forecasting (i.e., forecasting the aggregated power generation of an entire region and not limited to a single PV site or solar farm).

\subsection{Deep learning methods in solar forecasting}

The application of DNNs in solar forecasting has seen success in recent years due to their higher feature extraction capabilities when compared with traditional feed forward neural networks (FFNN). In particular, commonly used DNNs in solar forecasting include Long Short Term Memory Networks (LSTM), Gated Recurrent Unit Networks (GRU) and Convolutional Neural Networks (CNN). \citet{Abdel-Nasser2019AccurateLSTM-RNN} used LSTM networks trained with historical PV power generation data and \citet{Qing2018HourlyLSTM} used LSTM networks trained with weather forecasts of the same day to forecast the solar power. \citet{duPlessis2021Short-termBehaviour} used LSTM and GRU networks to forecast the power generation of a large PV plant at different levels of the PV system (i.e., forecasting the entire plant vs forecasting the generation at an inverter-level). The use of CNNs has become increasingly popular due to their deep features extraction capabilities compared to other DNNs \cite{Korkmaz2021SolarNet:Forecasting, Acikgoz2022AForecasting}. 1D and 2D CNNs are commonly used in the solar forecasting literature. Often with time series datasets 1D CNNs are used to capture the features in the time dimension. However, when using sky/ satellite imagery or when input features (e.g. past power generation lags or weather features) to the CNN are organised in a two dimensional representation, 2D CNNs are used in studies. For example, in \cite{Yu2020ImprovedForecast, Acikgoz2022AForecasting, Heo2020DigitalRegions, Zhang2018DeepNowcasting}, authors used 2D CNNs to forecast the solar power generation and \citet{Zang2020Day-aheadLearning} explored the application of two recent CNN architectures, Residual networks (ResNet) and Densely connected convolutional networks (DenseNet) to forecast the power generation of a solar PV plant and showed that ResNet and DenseNet outperform the use of conventional 2D CNNs. While CNNs have the capability to extract meaningful features for the learning task, LSTM and GRUs are suited for learning long term temporal patterns of the data. Therefore, the combination of LSTM and CNNs is also explored in solar forecasting studies. In, \cite{Wang2019PhotovoltaicNetwork, Kumari2021LongForecasting} authors showed that the combination of LSTMs and CNNs can outperform the use of a single network. Recently another class of CNNs known as dilated Temporal Convolutional Networks (TCN) were proposed to overcome the short-coming of capturing long-term temporal dependencies in traditional CNNs \cite{Lea2017TemporalDetection}. Subsequently, the application of TCNs were explored in solar forecasting studies. \citet{Lin2020TemporalForecasting} compared the performance of TCNs with LSTM and GRUs for solar forecasting and found that TCNs outperform LSTM and GRU networks. A similar work with TCNs was conducted in \cite{Mashlakov2021AssessingForecasting} where TCNs showed promising results in forecasting the solar power generation compared to other DNN counterparts.

\subsection{Regional solar power forecasting}
\label{sec:regional_relatedwork}
Forecasting the output of all distributed solar PV in an entire region can be challenging - depending on the region's size, some systems may be affected differently by environmental conditions such as cloud cover than others. Existing studies approach the problem of regional solar power forecasting in multiple ways depending on the availability of data. Although smart meter datasets are becoming increasingly available in many countries, obtaining such datasets can be difficult due to privacy concerns. Therefore, some works attempt to forecast regional solar power generation without using any power generation time series datasets. Typically these approaches use satellite imagery and/or meteorological data to estimate solar irradiation in the region of interest~\cite{Huertas-Tato2020ALearning,Deo2017ForecastingQueensland,Heo2020DigitalRegions}. Another set of studies consider power generation time series data from a subset of PV systems, often known as reference systems, to provide a regional forecast. In these approaches, upscaling of the power generation forecasts from the reference systems is required to provide forecasts for the whole region ~\cite{Koster2019Short-termLuxembourg,Lorenz2011RegionalIntegration,Fu2019ACriterion,Laevens2021AnStatistics,Saint-Drenan2019BayesianForecasting,Pierro2017Data-drivenData,Kim2021AInstallation}. However, the assumption made here is that the power generation data from reference systems are representative of the whole region of interest to provide a reliable forecast \cite{Lorenz2011RegionalIntegration}.

Furthermore, there exist studies that use the power generation time series datasets collected from all PV systems in the region of interest. However, these studies limit the use of input time series datasets to the forecasting method either by using (i) the aggregated power generation time series data (i.e, the solar power generation time series of the whole region); or (ii) by using time series datasets at a more granular level, such as rooftop PV level or a postcode level (i.e., individual time series datasets). The motivation for using the aggregated time series data is that it has similar characteristics (e.g., daily generation profile) to individual power generation datasets but is less sensitive to local weather conditions than individual power generation datasets \cite{Zhang2014ForecastingLevel}. Furthermore, using vast volume of weather datasets from different locations in the region may cause learning problems for some forecasting models \cite{DaSilvaFonsecaJunior2014RegionalAnalysis}. Therefore, univariate forecasting approaches are often followed when forecasting the regional solar power generation (i.e., forecasting only based on the historical information of the time series and not using exogenous data such as weather data). Studies \cite{Almaghrabi2021ForecastingApproach, Almaghrabi2021SpatiallyNetworks, Rana2020AProduction, Zhang2014ForecastingLevel} used methods such as Auto Regressive Integrated Moving Average (ARIMA), Least Squares Support Vector Machine (LS-SVM), Support Vector Regression (SVR), Random Forest (RF), FFNN, LSTM, 1D CNN and 1D CNN combined with LSTMs to forecast the regional solar power generation univariately. However, weather data (e.g. cloud cover, temperature) has a direct impact when forecasting the solar power generation. Therefore, to use weather data from different locations within the region, some works utilised dimensionality reduction techniques such as principle components analysis (PCA) to reduce the input weather features given to the forecasting models with the aggregated time series \cite{DaSilvaFonsecaJunior2014RegionalAnalysis, DaSilvaFonseca2014RegionalMethods}.

Individual power generation datasets are correlated with weather data collected at different locations within a region. Therefore, some studies first forecast the individual power generation time series with associated weather data and afterwards aggregate the forecasts to derive the regional solar power forecast \cite{DaSilvaFonsecaJunior2014RegionalAnalysis, Zhang2019AForecasting}. The work in \cite{DaSilvaFonsecaJunior2014RegionalAnalysis} found that for geographically large regions with varying weather conditions, building a forecast model for each individual power generation time series is more suitable due to the smoothing effect caused when adding the forecasts (i.e, overestimation of forecasts for one time series is compensated by the underestimation for the other time series). However, building a model for each individual power generation time series in a region may not be practical and computationally intensive for larger regions with hundreds or thousands of PV systems. To overcome this, \cite{Zhang2020AForecast, Yu2020ImprovedForecast} attempt to build a single forecasting model with DNNs trained using individual power generation and weather datasets to forecast the regional solar power generation. Although these works use DNNs that can learn from large datasets, only geographically smaller regions with a very small number of PV systems (e.g., 10 PV systems) are considered for regional forecasting. However, in a region of interest, there may be hundreds of distributed PV systems that are often dispersed across large geographical areas and affected by various weather conditions that needs to be accounted for to achieve an accurate regional forecast.

In this paper, we introduce a regional forecasting method that combines the advantages of the above two approaches that use either the aggregated time series or individual time series datasets. We model these time series together using deep learning to capture cross-series information that will facilitate an improved regional forecast while significantly reducing the number of forecasting models that are required. The regional distribution of the individual power generation datasets can be taken into account (i.e., similarities of the power generation profiles of nearby systems and dissimilarities of systems far apart), allowing diverse environmental conditions of different parts of the region to be accounted for. Furthermore, the smooth patterns associated with the aggregated data can also be captured by using the aggregated power generation data. With this work, we also explore the capability of DNNs to learn from a large volume of diverse time series datasets to make an accurate forecast.

There are two noteworthy points. Firstly, we provide an aggregated forecast for the systems that we have data for. However, in regions where power generation data is only available for a subset of systems (as discussed before), this forecast may be upscaled with existing upscaling approaches \cite{Bright2018ImprovedSystems}. It should also be noted that a full forecast for all systems is not always necessary, since for many purposes a forecast of a subset of systems can be sufficient. For example, suppose we can forecast that the aggregated output of 10\% of systems in a given region will drop by 30\% in one hour. In that case, it is reasonable to assume that this applies to the broader set of all systems. For system and network operators, a forecast of the relative increase or decrease of distributed generation is often sufficient to activate processes to mitigate problems. Therefore, even in regions where all data is not available, our approach can be used to provide relative forecasts accounting for the available power generation data. 

Secondly, when power generation data in a region of interest is represented in a time series hierarchy, another problem discussed in the literature is having incoherent forecasts across different levels of the time series hierarchy. This means that although the historical time series data is aggregate consistent, the generated forecasts are not (i.e., individual power generation forecasts do not add up to forecasts generated using the aggregated time series). This problem in the context of solar forecasting is discussed in \cite{Yang2017ReconcilingHierarchy}. The state-of-the-art approaches to address incoherent forecasts involve forecasting each time series separately and then reconciling the forecasts across the hierarchy. In this work, our goal is to provide an accurate regional (i.e., aggregated solar power generation forecast) and not forecast coherency. However, improved aggregated forecasts obtained from our approach can be used for reconciliation if the problem of interest is achieving forecast coherency.

\section{Methods}
\label{sec:method}

In the following, we first describe the advantage of jointly modelling an aggregated time series with individual time series data using a single network. Next, the main components of TCNs are explained. We then explain the proposed HTCNNs. Finally, two strategies on how HTCNNs can be applied to regional solar power forecasting are introduced.

\subsection{Modelling aggregated time series with individual time series datasets}

Let us consider $N$ individual time series $ts_1, ts_2, \dots, ts_N$ of length $r$ that forms an aggregated time series $ts_{AGG}$. $ts_j$ = $\{p_{j1}, p_{j2}, \dots p_{jr}\}$ represents an individual series $j$ and $p_{jr}$ reflects the value at time point r. Then, $ts_{AGG} = \{p_1, p_2, \dots p_r\}$ is such that for any time point $i$,

\begin{equation}
    p_i = \sum_{j=1}^N{p_{ji}}
\end{equation}

\begin{figure*}[]
    \centering
    \includegraphics[width=\textwidth]{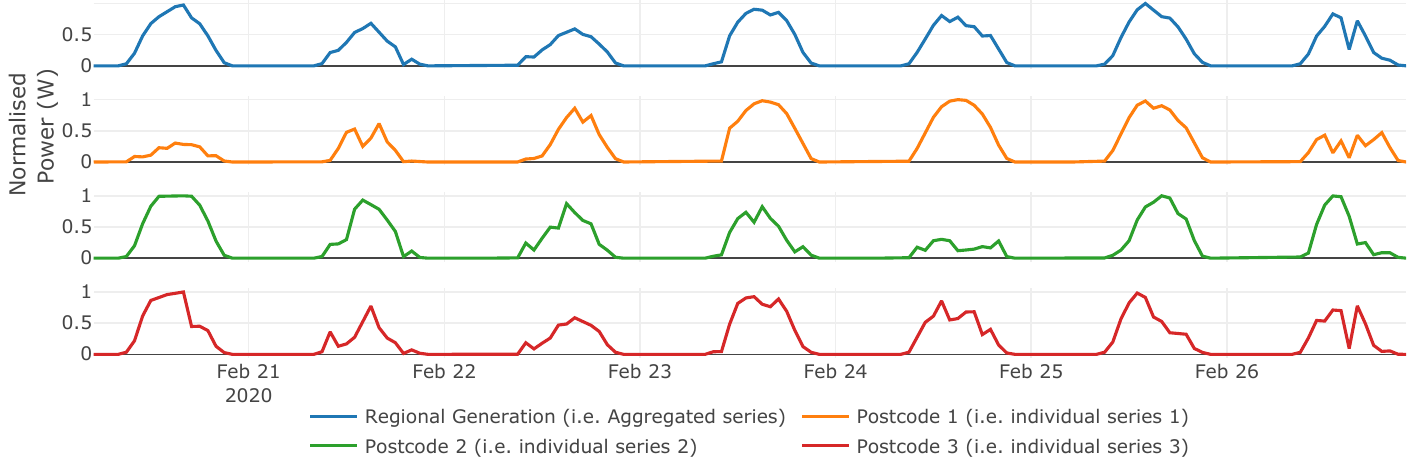}
    \caption{An example of normalised regional solar power generation and solar power generation at three different postcodes of this region.}
    \label{fig:solar_generation}
\end{figure*}

In a regional solar power forecasting setting, $ts_{AGG}$ is the regional solar power generation time series and $ts_1, ts_2, \dots, ts_N$ may represent the power generation time series of rooftop PV level or postcode level. This creates a time series hierarchy of two levels where the top level is created as an aggregate or sum of its bottom level time series. Recent studies \cite{Hollyman2021UnderstandingReconciliation} have discussed that each time series in such a hierarchy can have different characteristics that are useful in forecasting. Figure \ref{fig:solar_generation} shows one such example. Data at a top level ($ts_{AGG}$) has smooth patterns compared to data at the bottom level as it is an aggregated generation profile of hundreds of diverse time series datasets. In contrast, bottom level data has unique characteristics that are not present at an aggregate level and have exogenous features that can be used in forecasting. For example, local weather conditions (e.g., temperature, cloud cover) collected from multiple locations of the region can be used as exogenous features to forecast the individual power generation data as they are more affected by local weather conditions compared to the aggregate generation. To this end, in this work, we model $ts_{AGG}$ with $ts_1, ts_2, \dots, ts_N$ (and their exogenous features) time series data in a single network to capture features across all different time series to facilitate an improved forecast for $ts_{AGG}$. 

\subsection{Dilated Temporal Convolutional Neural Networks}
\label{sec:TCN}

TCNs are a type of CNN for sequence modelling tasks that are adapted from the WaveNet architecture \cite{Oord2016WaveNet:Audio} which was originally designed for speech synthesis tasks. TCNs were first introduced for action segmentation, and detection \cite{Lea2017TemporalDetection} and subsequently used for time forecasting tasks due to their ability to capture long-term temporal patterns in time series data \cite{Bai2018AnModeling, Lin2020TemporalForecasting}. 

\begin{figure*}[]
    \centering
    \includegraphics[width=0.8\textwidth]{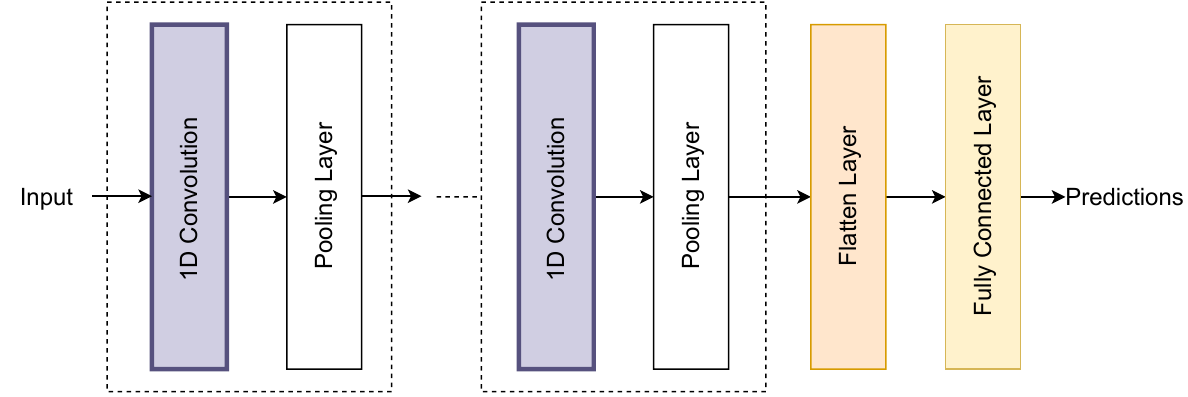}
    \caption{A traditional 1D convolutional neural network architecture for forecasting which consists of standard 1D convolutions and pooling layers. In addition, other layers such as dropout, normalization maybe included to prevent over fitting of the network.}
    \label{fig:traditionalCNN}
\end{figure*}

Figure \ref{fig:traditionalCNN} shows a traditional 1D CNN structure which consists of 1D convolution and pooling layers. As shown in Figure \ref{fig:convExample}, the 1D convolution layers create a feature representation of the given input by convolving (dot product) a filter (weight vector) across the input. The pooling layers create a downsampled feature map (i.e., reduces the dimensionality by summarising the features). This helps to capture dominated features and reduce the number of parameters the network needs to learn. On the contrary, TCNs do not have pooling layers but instead use ``dilated" convolution where the filter is applied over an area larger than the filter size by skipping intermediate values in the input by a certain number of steps (i.e., dilation/ dilation rate). An example of dilated convolution with a dilation rate of 2 is shown in Figure \ref{fig:dilatedconvExample}. 

\begin{figure}[]
    \centering
    \includegraphics[width=\textwidth]{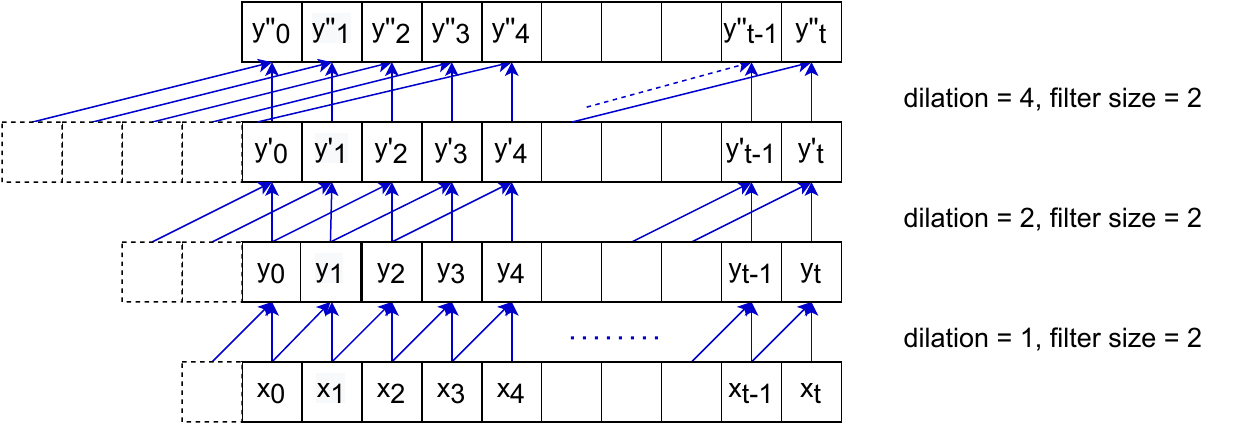}
    \caption{Visualisation of a stack of dilated causal convolutions (with dilation rates 1, 2, 4 and padding where zero padding is shown with dash squares) on an input sequence $x_0, x_1, x_2,\dots,x_{t-1},x_t$. The blue lines show a filter with a filter size of 2. The output sequences generated after applying the filters are shown as $y$/ $y'$/ $y''$. }
    \label{fig:dilatedConv}
\end{figure}

\begin{figure}[]
    \centering
    \includegraphics[width=0.5\textwidth]{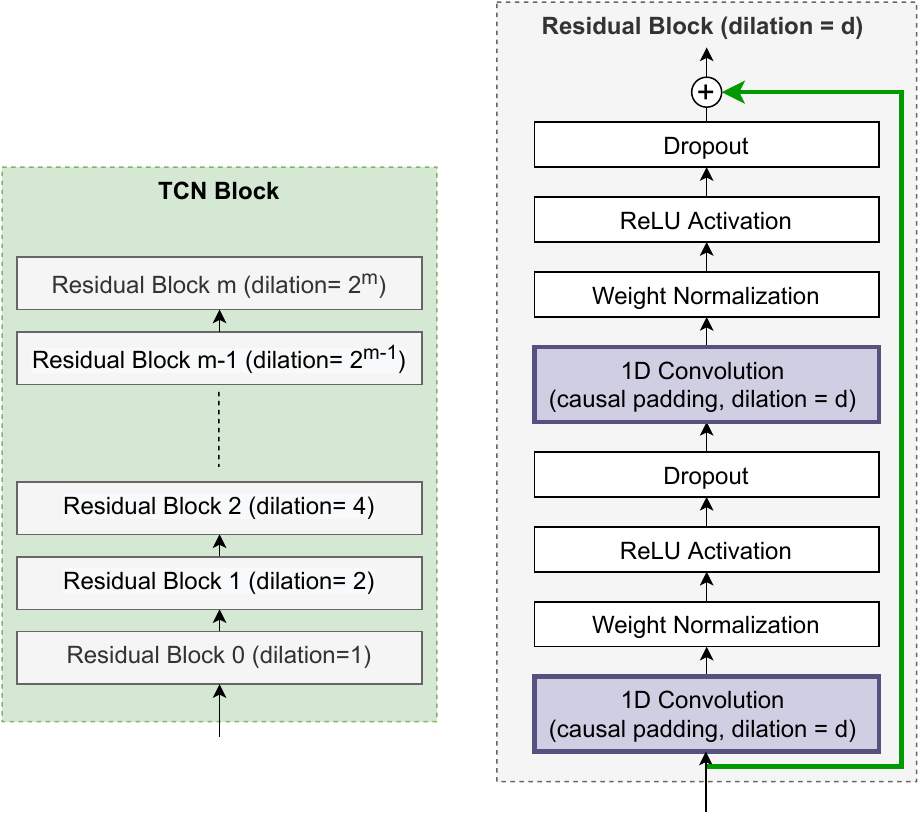}
    \caption{Main components of the Dilated TCNs. Left: shows a TCN block which is composed of $m+1$ Residual Blocks, with $2^d$ dilation rate for a Residual block $d$. The output of Residual block $d$ will be the input to Residual block $d+1$. Right: shows the components of a single Residual Block, $d$, that consists of two 1D convolution layers with a $d$ dilation rate and causal padding, Weight normalizations, ReLU activation, Drop out layers and a skip connection (shown in green) from the input to the output of the last Dropout layer.}
    \label{fig:temporalConv}
    % \vspace{-4}
\end{figure}

 In TCNs, long-term temporal patterns of the data are captured through stacking several dilated 1D convolutions as shown in Figure \ref{fig:dilatedConv}, where the dilation rate increases with the depth of the network. As the dilation increases, the filtered feature output can capture a wider range of the input, thus increasing the network's receptive field (i.e., how far to the history can be seen). Figure \ref{fig:temporalConv} shows the main components of the TCN and how the 1D dilated convolutions are stacked within the network. A TCN with $m+1$ layers will have $m+1$ residual blocks with a $2^i$ dilation for a residual block $i$. The power of 2 dilation increase with the depth of the network allows the TCN to capture long-term temporal patterns. Each residual block will have two 1D convolution layers with the respective dilation of the residual block. Furthermore, weight normalization and dropout layers are added to prevent overfitting of the network. Each residual block also consists of skip connections (shown in green) where the input to a residual block is added back to its output. The purpose of the skip connections in DNNs is to facilitate the training of deeper networks by allowing the network layers to learn modifications to the identity mapping (i.e., learn the change to the input than learning the entire transformation of the input) \cite{he2016deep}.

In addition to the ability of TCNs to capture long-term temporal features of the data, they are also faster to train than LSTM models \cite{Bai2018AnModeling, Lea2017TemporalDetection}, which is important when training with a larger volume of diverse data. Therefore, in this work, TCNs blocks (shown in Figure \ref{fig:temporalConv}) are used within the HTCNNs architectures to capture features from the time series datasets.

\subsection{Proposed Hierarchical Temporal Convolutional Neural Networks}

In this section, the main components of HTCNNs are explained. The motivation for both HTCNNs is to model $ts_{AGG}$ with $ts_1, ts_2, \dots, ts_N$ and their exogenous features in a single network to learn across diverse time series datasets. However, two architectures are different in terms of their design to capture features from the aggregated and individual data. Figure \ref{fig:HTCNN1}, \ref{fig:HTCNN2} shows the two architectures: HTCNN Architecture 1 (A1) and HTCNN Architecture 2 (A2) respectively. Similar colors are used to indicate similar layers in the two architectures. 

Commonly, to forecast a solar power generation time series, historical power generation data and weather forecasts are provided as inputs to a forecasting model. Thus, for any solar power generation time series ($ts$), to forecast a horizon of $h$, we denote the associated input feature matrix of size $t\times f$ as follows, where $t$ denotes the number of rows, and $f$ denotes the number of columns. $f^x$ denotes a feature $x$ (e.g. historical power generation, weather forecasts) and $m$ denotes the number of features.

\begin{equation}
\left[
\begin{array}{cccc}
f^1_{1} & f^2_{1} & \dots & f^m_{1} \\
\vdots & \vdots & \ddots & \vdots \\
f^1_{t} & f^2_{t} & \dots & f^m_{t}
\end{array}
\right]
\label{eq:feature_matrix}
\end{equation}

\begin{figure}[!h]
    \centering
    \includegraphics[width=0.6\textwidth]{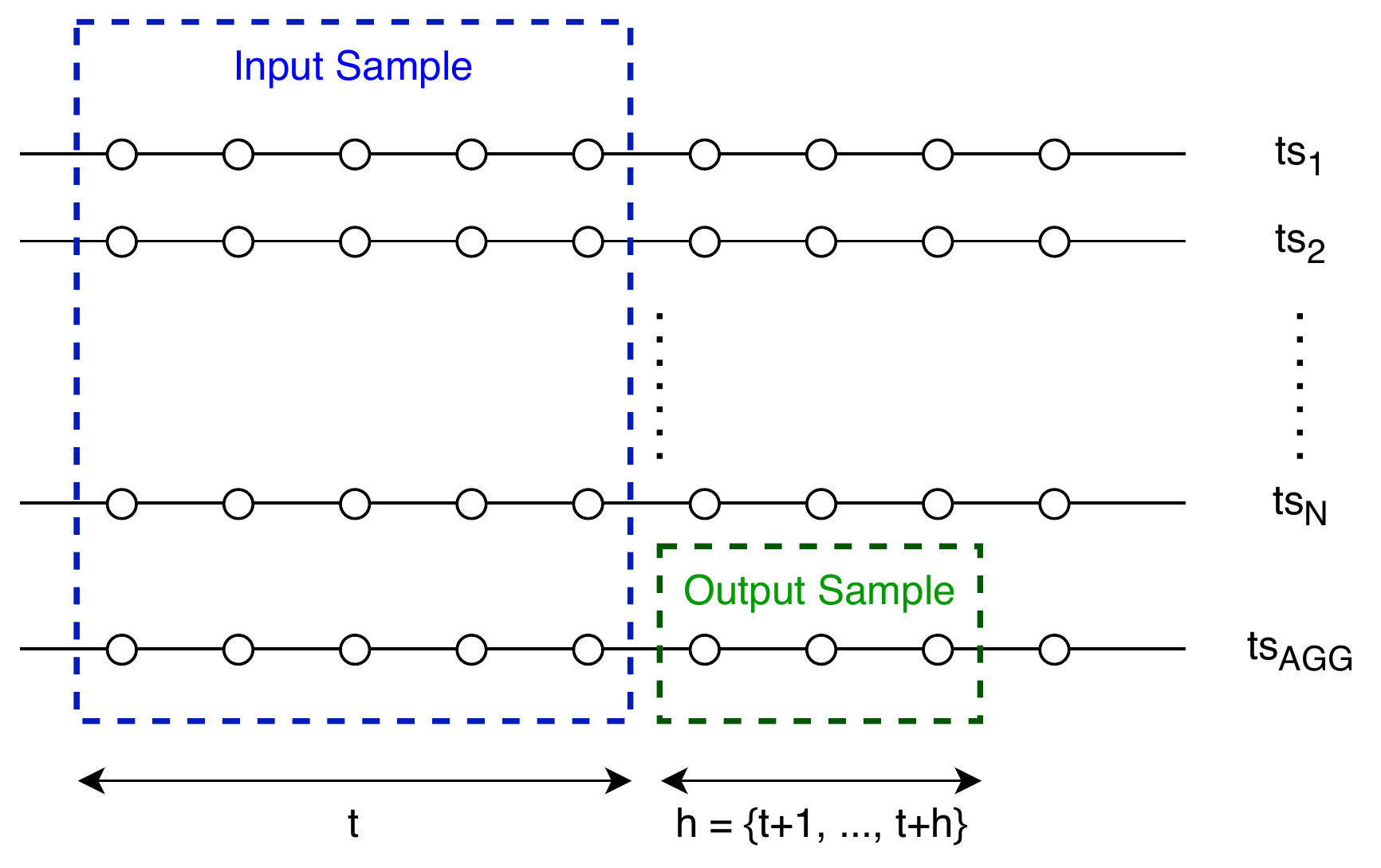}
     \caption{An overview of the input and output samples of time series data used to train the proposed HTCNNs.}
    \label{fig:ts_samples}
\end{figure}

In this work, we use the following input and output data samples shown in Figure \ref{fig:ts_samples} to train the HTCNNs. To forecast $h$ steps ahead for $ts_{AGG}$, the input samples corresponds to input feature matrices from $ts_1, ts_2, \dots, ts_N$ and $ts_{AGG}$. The individual time series and aggregated time series shown in Figure \ref{fig:HTCNN1}, \ref{fig:HTCNN2} refers to these different input feature matrices.

\subsubsection{HTCNN Architecture 1}
\label{subsec:HTCNNA1}

\begin{figure*}[]
    \centering
    \includegraphics[width=0.9\textwidth]{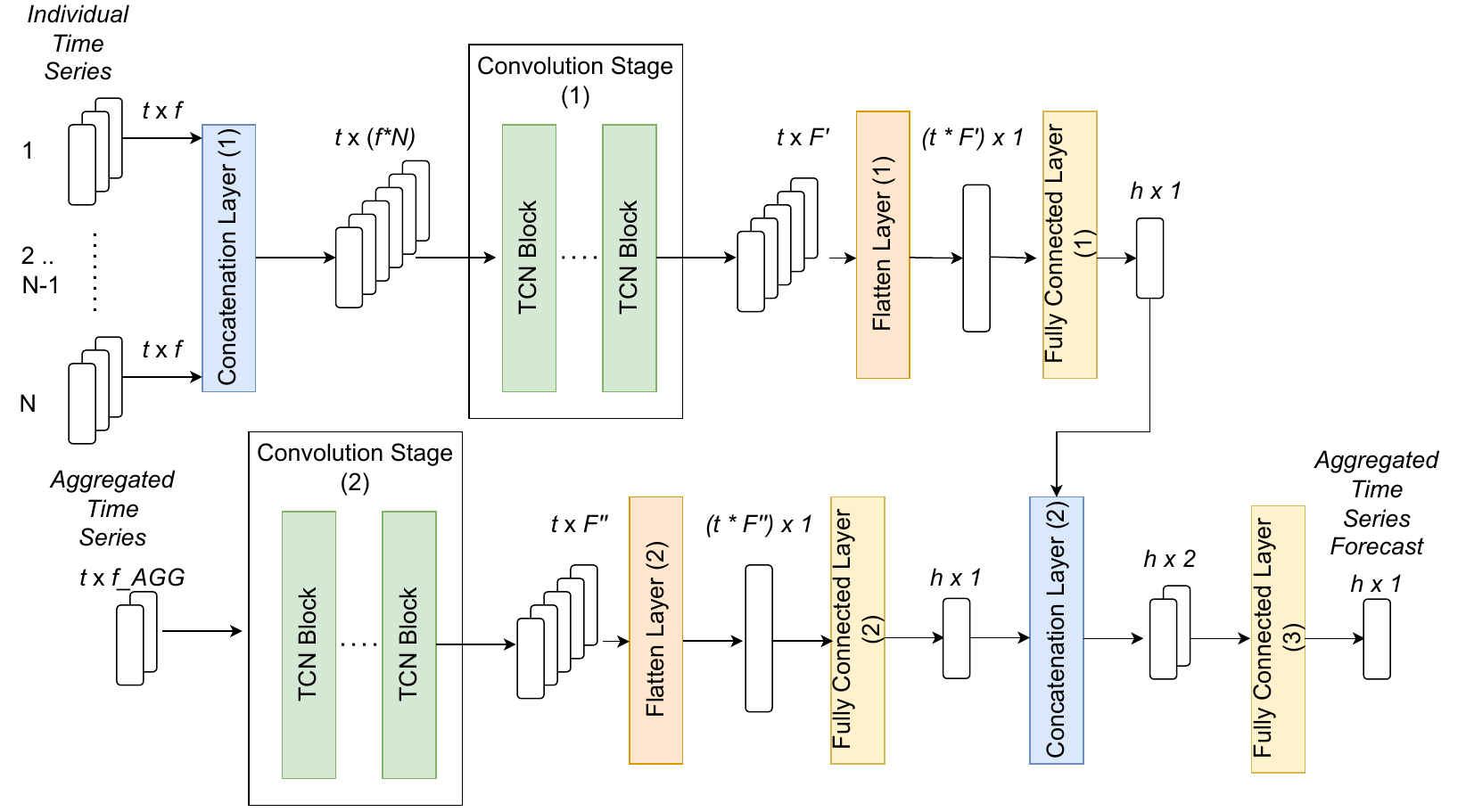}
     \caption{An overview of the proposed HTCNN Architecture 1. $N$ is the number of individual time series, $t$ is the number of time steps in a single input sample, $f$ is the number of features available for an input sample in the individual data, $f\_AGG$ is the number of features available for an input sample of the aggregated data, $F', F''$ corresponds to the number of convolution filters in the convolution layers, $h$ is the desired forecast horizon.}
    \label{fig:HTCNN1}
\end{figure*}

Let us consider time series structured in a hierarchy with two levels (e.g., considering the regional level and postcode level in Figure \ref{fig:heirarchy}). The data can be divided into two main categories, (i) data at the bottom level or individual and (ii) data at the top level or the aggregated data. As can be seen in Figure \ref{fig:HTCNN1}, there are two separate paths in the HTCNN A1 to learn from these two categories of data. HTCNN A1 has four types of components: Concatenation Layers, Convolution Stages, Flatten Layers and Fully Connected Layers. When there is more than one component of the same type, we use the (1), (2), \dots numbering to distinguish them (e.g., there are two Convolution Stages and is denoted as Convolution Stage (1), Convolution Stage (2)).
\vspace{3mm}
\newline
\textbf{Concatenation Layers}: Concatenation layer (1) combines the individual time series datasets before passing to the convolution stage. Concatenation layer (2) combines the outputs of the fully connected layers (1) and (2) and pass to the last fully connected layer that forecasts the aggregated time series. The concatenation in both layers happens only in the $y$ dimension (i.e., feature space). For example, concatenating two feature matrices $l$ and $m$ with dimensions $t\times k_{1}$ and $t \times k_{2}$ result in a feature matrix of $t\times (k_{1}+k_{2})$.
\vspace{3mm}
\newline
\textbf{Convolution Stage}: There are two convolution stages to capture features from the two different categories of data. Convolution Stage (1) capture features from the individual time series datasets, while Convolution Stage (2) capture features from the aggregated time series. The number of TCN blocks and the number of filters $F'$ and $F''$ in convolution stages are hyper-parameters that need to be tuned for a given dataset. The $x$ dimension of the 2D input data does not change after a convolution stage (e.g. $t\times(f*N)$ will become $t\times F'$ after convolution stage (1)) as we use TCN blocks with dilated causal convolution as discussed in Section \ref{sec:TCN}. 
\vspace{3mm}
\newline
\textbf{Flatten Layers}: The flatten layers are used to flatten the 2D feature maps generated by the convolution stages into a 1D vector. Flatten layer (1) flattens the feature maps generated for the individual time series, while flatten layer (2) flattens the feature maps generated for the aggregated series.
\vspace{3mm}
\newline
\textbf{Fully Connected Layers}: The purpose of the fully connected layers (1) and (2) is to learn the non-linear mappings of the features extracted from the convolution stages (1) and (2), respectively. Fully connected layer (3) compute the non-linear relationships of the predictions made by fully connected layers (1) and (2). When the desired forecast horizon is $h$, fully connected layer (3) outputs a $h\times1$ vector corresponding to the predictions for each time step. We have experimentally determined the output vector size of fully connected layers (1) and (2) to be the same as $h$. However, this vector size can also be considered as a hyper-parameter.

\subsubsection{HTCNN Architecture 2}
\label{subsec:HTCNNA2}
\begin{figure*}[]
    \centering
    \includegraphics[width=\textwidth]{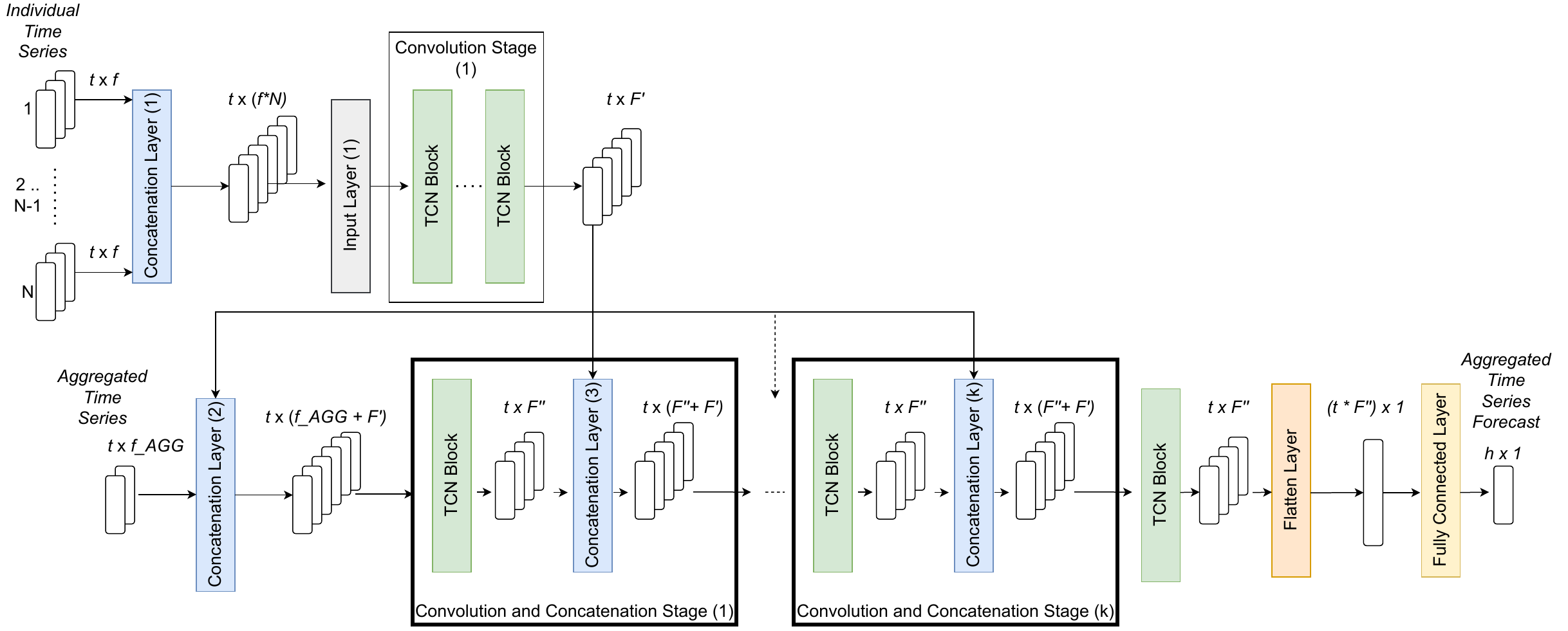}
    \caption{An overview of the proposed HTCNN Architecture 2. $N$ is the number of individual time series, $t$ is the number of time steps in a single input sample, $f$ is the number of features available for an input sample in the individual data, $f\_AGG$ is the number of features available for an input sample of the aggregated data, $F', F''$ corresponds to the number of convolution filters in the convolution layers, $h$ is the desired forecast horizon.}
    \label{fig:HTCNN2}
\end{figure*}

HTCNN A2 has five types of components: Concatenation Layers, Convolution Stages, Convolution and Concatenation Stages, Flatten Layers and Fully Connected Layers. Similar to HTCNN A1, when there is more than one component of the same type, we use the (1), (2), \dots, numbering to distinguish them.
\vspace{3mm}
\newline
\textbf{Concatenation Layers}: Concatenation layer (1) is the same as HTCNN A1. Concatenation layer (2) combine the convolution features extracted from individual time series datasets with the input features of the aggregated data. This concatenation is done for the bottom branch of the network to learn from both aggregated data and features captured from the individual time series datasets.
\vspace{3mm}
\newline
\textbf{Convolution Stage}:  Convolution stage (1) consists of TCN blocks to capture features from the individual time series data. Often a large number of dataset are available at an individual time series level. While our interest is to forecast the aggregated time series ($ts_{AGG}$), it is also essential features from individual data that can help in forecasting $ts_{AGG}$. Therefore, the individual data is first sent through a convolution stage to capture such features. The number of filters ($F'$) and TCN blocks in the convolution stage is a hyper-parameter that needs to be tuned.
\vspace{3mm}
\newline
\textbf{Concatenation and Convolution Stages}: As can be seen in Figure \ref{fig:HTCNN2}, there are $k$ number of concatenation and convolution stages. $k$ is a hyper-parameter that needs to be tuned. In a concatenation and convolution stage, the input data is first sent through a TCN block to extract features from the input, followed by a concatenation layer. The concatenation layer will combine the feature output of the TCN block with the features extracted from the individual data. Thus, a concatenation and convolution stage $k$ ($k >1$) receives combined feature maps from Convolution Stage (1) and TCN block in Concatenation and Convolution Stage (k-1). This approach of convolution and concatenation was inspired by the DenseNet architecture \cite{Huang2017DenselyNetworks} in which feature reuse and propagation in a DNNs have shown promising results. However, in DenseNet, every layer receives feature maps from all proceeding layers. On the contrary, we reuse features extracted from individual time series data because a large volume of exogenous features at an individual level needs to be accounted in regional forecasting (e.g., weather features are available at an individual time series level). It is also possible for the concatenation to only happen at the beginning, end or in some intermediate layers in the network. However, through experiments, we found that concatenating features after every convolution layer results in better forecast accuracy.
\vspace{3mm}
\newline
\textbf{Flatten Layer and Fully Connected Layer}: The flatten layer is used to flatten the 2D feature maps into a 1D vector. Similar to HTCNN A1, the last fully connected layer computes the non-linear relationships of the extracted features and the desired output.

\subsection{Proposed Regional Solar Power Forecasting Strategies}

This section introduces two strategies to adapt HTCNN A1 and A2 for regional solar power forecasting. The strategies are independent of whether the network is A1 or A2. Therefore, in the following section, the network is only referred to as HTCNN. 

\begin{figure}[!h]
     \centering
     \subcaptionbox{The direct regional forecast strategy with HTCNN. \label{fig:strategy1}} %
    [.4\textwidth]{\includegraphics[width=\linewidth]{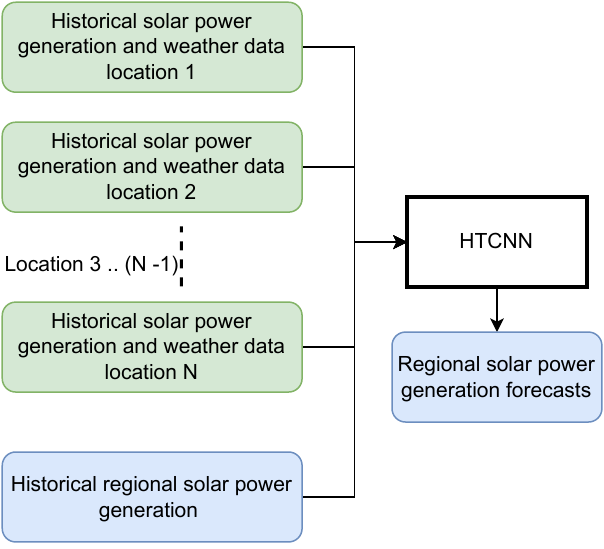}}
     \hspace{10pt}
    \subcaptionbox{Sub region aggregation forecast strategy with HTCNN. First, the sub-region power generation is forecasted separately with a HTCNN. Next, these forecasts are added to derive the regional solar power generation. \label{fig:strategy2}}
      [.5\textwidth]{\includegraphics[width=\linewidth]{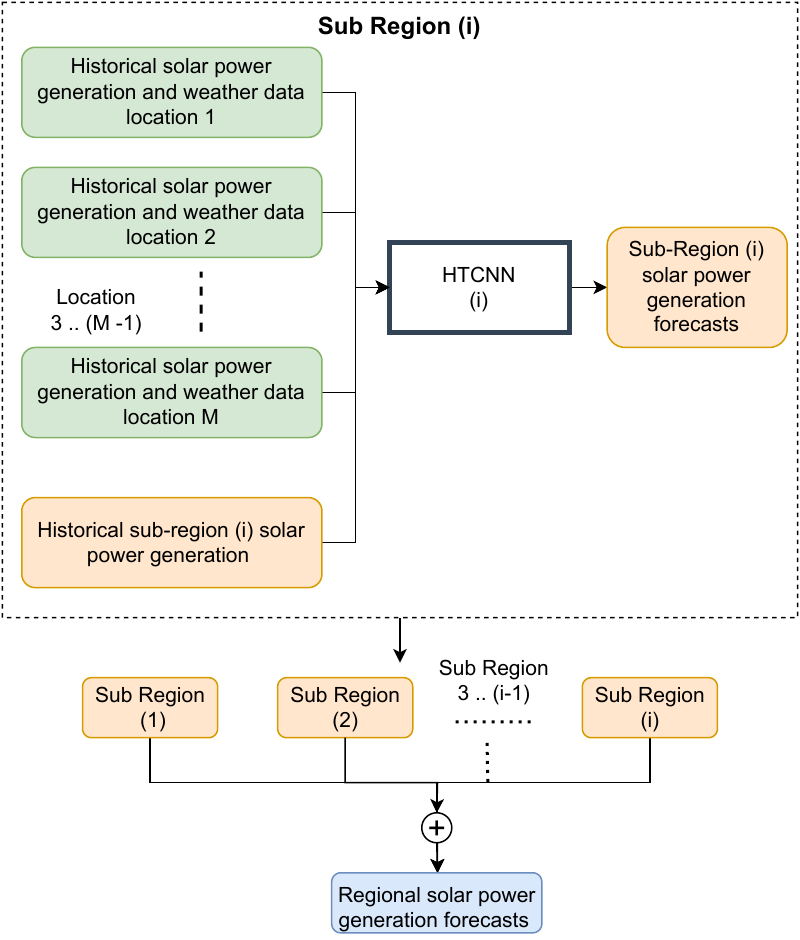}}
    %  \begin{subfigure}[]{0.4\textwidth}
    %      \centering
    %      \includegraphics[width=\textwidth]{figures/Strategy1.1.pdf}
    % \caption{The direct regional forecast strategy with HTCNN.}
    % \label{fig:strategy1}
    %  \end{subfigure}
    %  \hfill
    %  \begin{subfigure}[]{0.5\textwidth}
    %      \centering
    %      \includegraphics[width=\textwidth]{figures/Strategy2.1.pdf}
    %     \caption{Sub region aggregation forecast strategy with HTCNN. First, the sub-region power generation is forecasted separately with a HTCNN. Next, these forecasts are added to derive the regional solar power generation.}
    %     \label{fig:strategy2}
    %  \end{subfigure}
        \caption{HTCNN-based regional forecasting strategies.}
        \label{fig:strategies}
\end{figure}

\subsubsection{Direct Regional Forecast with HTCNN}
\label{sec:direct_regional}

In this regional forecasting strategy, the solar power generation of the whole region of interest is directly forecasted (i.e., the output of the HTCNN) is the desired regional solar forecast. Therefore, only one HTCNN is required to forecast an entire region. Figure \ref{fig:strategy1} shows an overview of the direct forecast strategy with HTCNN. Given the historical power generation, weather datasets from all the locations in the region and the historical power generation of the regional solar generation, a single HTCNN is trained to forecast regional solar power generation.

\subsubsection{Sub Region Aggregation Forecast with HTCNN}
\label{sec:subregion_agg}

\begin{figure}[]
    \centering
    \includegraphics[width=\textwidth]{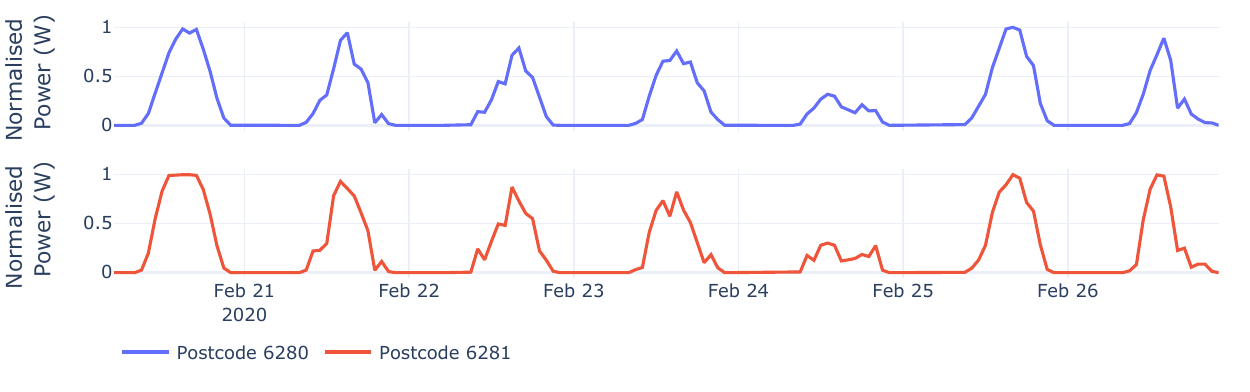}
    \caption{Normalised power generation of two nearby postal codes in Western Australia.}
    \label{fig:nearby_pcs}
\end{figure}

Power generation datasets coming from nearby locations have closely related generation profiles as they may be affected by similar weather conditions. An example of this can be seen in Figure \ref{fig:nearby_pcs}. Therefore, in the sub-region based division, a single HTCNN is trained with data that are closely related to each other. This is achieved with the following approach:
\begin{enumerate}
    \item Divide the locations of the region into sub-regions based on the closest weather collection point to a given location
    \item Forecast each sub-region with a HTCNN specific to the sub-region
    \item Aggregate (i.e., add) the forecast from all sub-regions to derive at the regional forecast
\end{enumerate}

Figure \ref{fig:strategy2} shows an overview of this strategy. With the sub-region based approach, there are (i) multiple HTCNNs and (ii) additional solar power generation time series for each sub-region which is an aggregate series of the individual power generation data belonging to a sub-region. Each HTCNN is trained to forecast the aggregated sub-region generation with the historical individual power generation, weather data from all the locations in the sub-region and the historical aggregated power generation of the sub-region as input.

\section{Case study in Western Australia}
\label{sec:case_study}

\subsection{Data}
\label{sec:data}

For the application of regional solar power forecasting, we consider the South West Interconnected System (SWIS) which is the largest electricity grid out of the two grids in Western Australia (WA), where one in three households connected to SWIS have rooftop solar installed \cite{operator2019integrating, Lu201790100Australia}. Rooftop solar is the primary generation source of SWIS and the Australian Energy Market Operator (AEMO) recently reported that rooftop solar in SWIS is expected to grow to provide 40\% of the total generation by 2030 \cite{operator2021integrating}. SWIS is a sparse grid that covers an area of approximately 261,000 $km^2$ and is geographically isolated from other electricity grids in Australia. Annually, it provides approximately 18 million megawatt hours of electricity to more than 1.1 million businesses, and households \cite{operator2019integrating}.

\begin{figure*}[!h]
    \includegraphics[width=\textwidth]{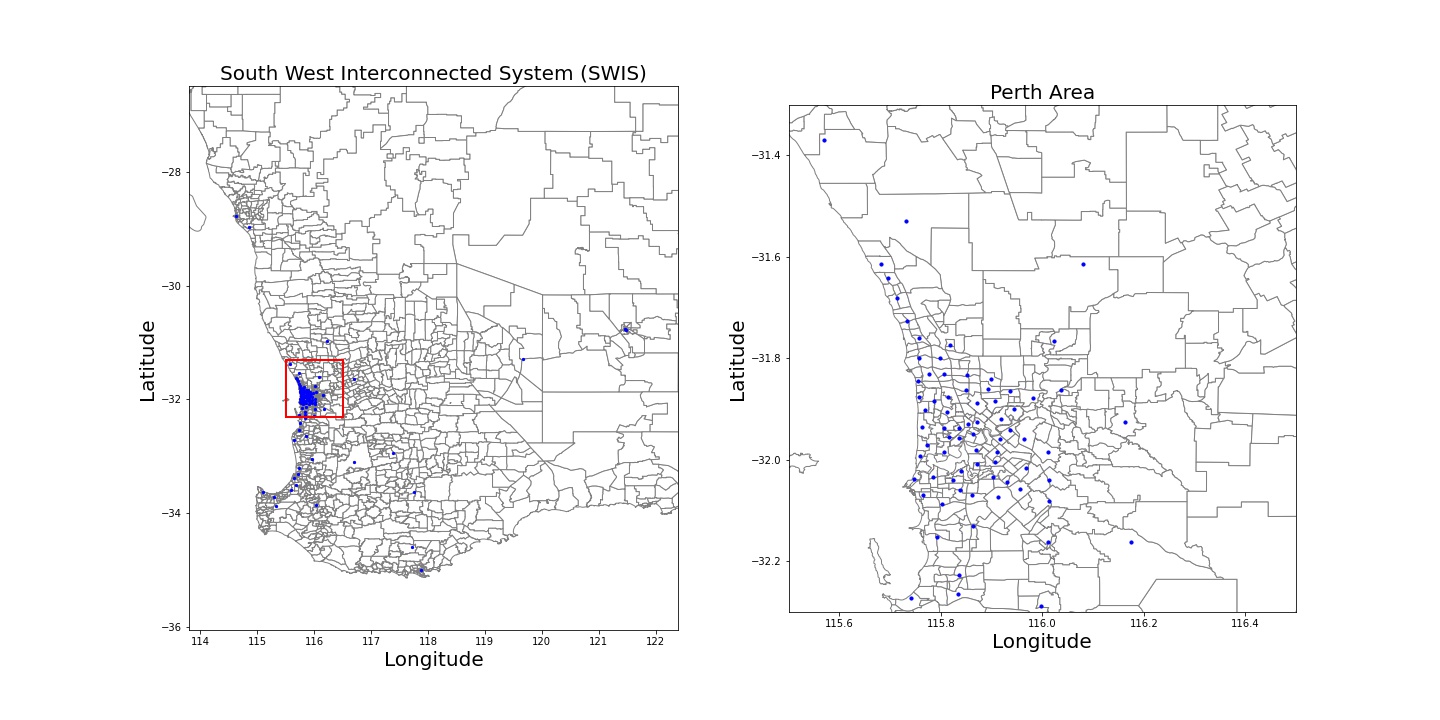}
    \caption{Left: South West Interconnected System (SWIS) of Western Australia, Right: A zoomed-in view of the Perth Area (shown with a red rectangle) in Western Australia shown. Blue dots represent the postcode centers.}
    \label{fig:swis_map}
\end{figure*}

Power generation time series are provided by Solar Analytics\footnote{www.solaranalytics.com.au}. This dataset comprises information from 241 rooftop solar PV systems within SWIS. The data covers the period from 13-Feb-2020 to 28-Feb-2021, with a time resolution of 1 hour. Each PV system is associated with a postcode, which provides an approximation of the location of the system. These PV systems are distributed across 101 postcodes in Western Australia, as shown in Figure \ref{fig:swis_map}. The figure shows the geographical dispersion of the PV systems, with a notable concentration around the area of Perth.

\begin{figure*}[!h]
    \centering
    \includegraphics[width=\textwidth]{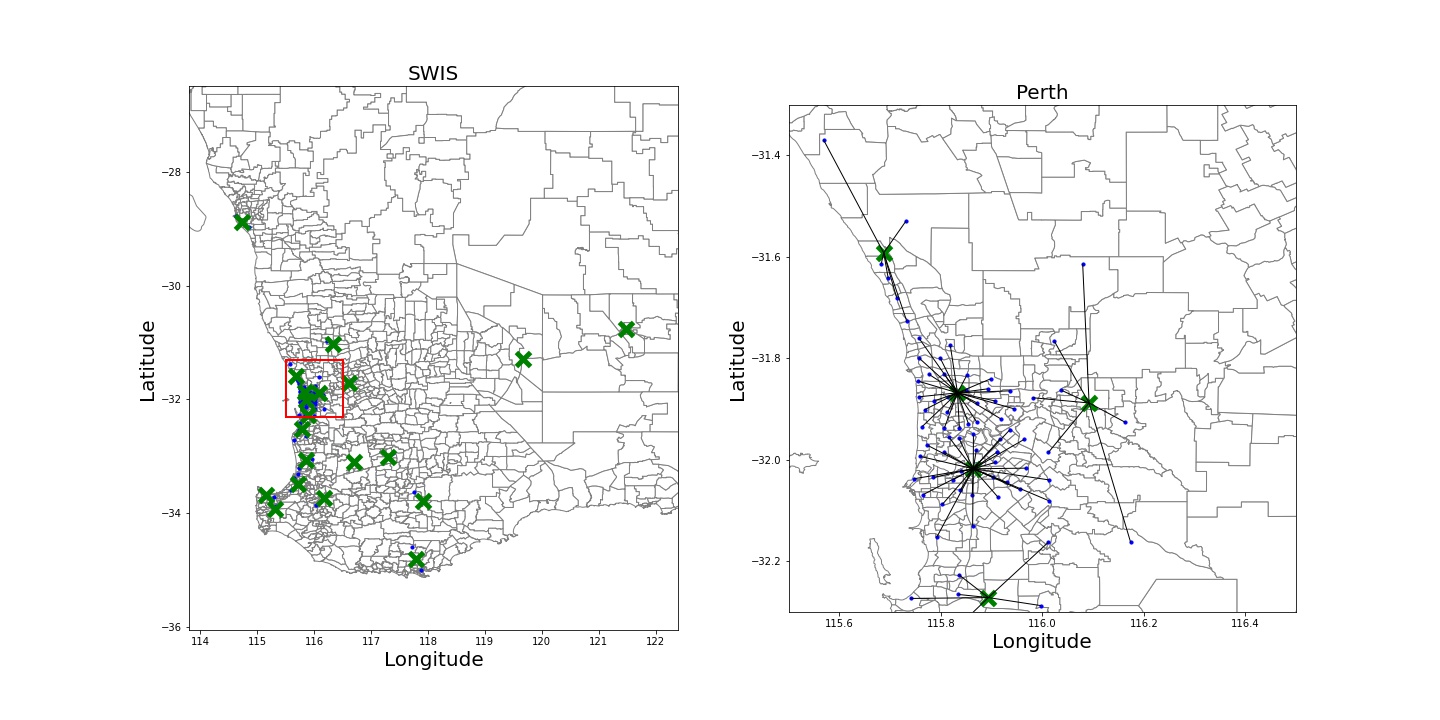}
    \caption{Left: Green ``X" shows the 20 weather collection locations, Blue dots shows the postcode centers in the South West Interconnected System (SWIS) of Western Australia, Right: A zoomed in area of the Perth Region, the black lines shows the associated weather collection locations for a particular postcode.}
    \label{fig:weather_collection_points}
\end{figure*}

Weather data given as exogenous features during the forecasting process is extracted from the Dark Sky online API \cite{darkSky}. Seven weather related features are considered, namely wind speed, temperature, UV index, cloud cover, humidity, pressure and dew point. To extract weather features from suitable weather collection locations, we apply K-means clustering using the latitude and longitude corresponding to the centroid of the geographical area of the postcode, resulting in 20 unique clusters. The cluster centroid coordinates (as illustrated in Figure \ref{fig:weather_collection_points}) are then used to collect weather from Dark Sky.

\subsection{Forecasting the Solar Power Generation of SWIS}

In this work, we generate a day-ahead forecast for SWIS at an hourly resolution using the proposed methods in Section \ref{sec:method} and benchmark methods in Section \ref{sec:baselines}. For each day forecasts are generated from 5 am to 10 pm of the same day, resulting in a forecast horizon ($h$) of 18 hours (i.e., 18 time points). Periods before 5 am and after 10 pm of the same day are not considered when forecasting as power generation values during these periods are often zeros across all locations due to low solar irradiance during early morning and night. 

SWIS has 101 postcodes and the power generation time series of each postcode location corresponds to $ts_1, ts_2, \dots, ts_N$ (Section \ref{sec:method}) where $N$ is 101 and $ts_{AGG}$ is the aggregated power generation of SWIS. For each postcode, weather data from the closest weather collection location to the postcode from the 20 weather locations based on the prior clustering (as illustrated in Figure \ref{fig:weather_collection_points}) is used as the related weather features of the postcode. According to the notation introduced in Section \ref{sec:method}, the input feature matrix size $t \times f$ for each postcode becomes $18 \times 14$. This refers to the past 7 days of historical power generation values and forecasting day's weather data (corresponding to the 7 weather features). For the regional generation time series of SWIS, the input feature matrix ($t\times f_{AGG}$) is of size $18 \times 7$, which denotes the past 7 days of historical power generation values. 

In the direct forecast strategy for SWIS, the above described values remain the same. However, in the sub-region based forecast strategy, there are 20 sub-regions as there are 20 weather data collection locations in SWIS. Therefore, in a single HTCNN, $ts_1, ts_2, \dots, ts_N$ reflect the power generation time series of the postcodes that are mapped to the particular sub-region, $N$ denotes the number of mapped postcodes and $ts_{AGG}$ reflects the aggregated sub-region power generation time series.

\subsubsection{Baseline Forecasting Models and Strategies for Comparison}
\label{sec:baselines}

We implement five state-of-the-art forecasting methods to evaluate the proposed HTCNNs in comparison to existing deep learning and classical time series-based methods that are well-established and extensively studied in the solar power forecasting literature \cite{Erdener2022AForecasting, Ahmed2020AOptimization, Wang2019ANetwork}. These methods include Seasonal Naive (SN) model, Seasonal Autoregressive Integrated Moving Average model (SARIMA) with and without exogenous features and three widely used DNNs: TCN, which is a building block of the proposed HTCNN architecture, 1D CNN, and LSTM. The 1D CNN and TCN are implemented as discussed in Section \ref{sec:method}. In the following, we describe SN, SARIMA and LSTM.
\vspace{3mm}
\newline
\textbf{SN:} SN (also known as Periodic Persistence) is a naive model that considers the forecasts are the same as the last observed power generation values. For example, for a day-ahead solar power forecast, this method considers the power generation of the forecasting day is similar to the power generation of the previous day. Although this is a simple method, SN is commonly used as a benchmark approach in many studies in the literature as it has shown to outperform other forecasting methods, particularly in short-term forecast horizons where future solar generation values are closely related to past generation values \cite{Trapero2015Short-termRegression, Barbieri2017VeryReview}. 
\vspace{3mm}
\newline
\textbf{SARIMA:} SARIMA is the seasonal variation of the ARIMA method. In an ARIMA \textit{(p,q,d)} model, as shown in Equation \ref{eq:arima}, the forecasting variable of interest is represented as a linear combination of the past values (autoregressive component) and past forecast errors $\varepsilon$ (moving average component). The three variables $p, q, d \in\mathbb{Z}$ reflect the order of the autoregressive, moving average and differencing components. The series after differencing is shown as $y'_{t}$. $\phi, \theta$ are coefficient and $c$ is a constant.

\begin{equation}
  y'_{t} = c + \phi_{1}y'_{t-1} + \cdots + \phi_{p}y'_{t-p}
     + \theta_{1}\varepsilon_{t-1} + \cdots + \theta_{q}\varepsilon_{t-q} + \varepsilon_{t}
\label{eq:arima}
\end{equation}

A SARIMA model includes an additional component $(P,Q,D)_m$ to represent the seasonal component, where $m$ reflects the seasonality (e.g., for a solar power generation time series with 18 time points per day, $m$ is 18 accounting for the daily seasonality). When exogenous features are modelled with a SARIMA model, it is known as SARIMAX.
\vspace{3mm}
\newline
\textbf{LSTM:} LSTM is a type of DNN that can capture short and long-term temporal dependencies in sequential data and therefore is widely used in sequence prediction problems, including forecasting studies \cite{Hewamalage2021RecurrentDirections}. LSTMs have a mechanism named ``gating" that controls the flow of information within the network by learning which data needs to be preserved or forgotten (a detailed description of the gating mechanisms in a LSTM network is provided in \ref{appendixB}). There are various LSTM architectures that are used in forecasting studies (e.g., stacked, encoder-decoder). In this work, we implemented a widely used architecture known as stacked LSTM \cite{Hewamalage2021RecurrentDirections} where multiple LSTM layers are stacked on top of each other. The first LSTM layer receives the input data. For all other LSTM layers, the output of the previous LSTM layer is fed as the input to the next immediate LSTM layer. The final LSTM layer is followed by a fully connected layer to generate the forecasts.\vspace{3mm}

These baseline forecasting methods are used to forecast the regional solar generation of SWIS using two existing regional forecasting strategies (discussed in \ref{sec:regional_relatedwork}): (i) using only the aggregated regional level time series (similar to work in \cite{Almaghrabi2021SpatiallyNetworks, Rana2020AProduction, Zhang2014ForecastingLevel}) and (ii) using the postcode level time series and associated weather data to forecast each time series independently and then aggregating the forecasts (similar to bottom-up approaches in \cite{DaSilvaFonsecaJunior2014RegionalAnalysis, DaSilvaFonseca2014RegionalMethods, Yang2017ReconcilingHierarchy}). Using only the aggregated time series requires a single forecasting model to forecast the regional solar generation, while using postcode level time series requires one forecasting model per postcode resulting in 101 forecasting models.

% Furthermore, for the TCN-based forecasting method, we evaluate the impact of reducing the 101 forecasting models using a similar sub-region-based approach proposed in \ref{sec:subregion_agg}. In this approach, a TCN is implemented for each sub-region in SWIS where each network is trained using postcode-level time series and weather data belonging to the sub-region to forecast the power generation of all post codes in the sub-region.

In addition, we evaluate the impact of reducing the number of forecasting models required when using the postcode level time series. To assess this impact, we implement TCN as a global forecasting model \cite{hewamalage2022global} in which the network is trained using all postcode level time series in a sub-region to forecast the power generation for all postcodes within that sub-region using one network. Sub-regions are a group of postcodes based on the closest weather collection location as discussed in Section \ref{sec:subregion_agg}. To derive the regional forecast, all postcode level forecasts are added.

% (where a model simultaneously learns from many time series instead of a single time series) for each sub-region (where sub-regions are a group of postcodes based on the closest weather collection location as discussed in Section \ref{sec:subregion_agg}). Each TCN network in a sub-region is trained using power generation time series and weather data specific to the sub-region's postcodes, enabling to forecast the power generation for all postcodes within a sub-region using one network. Finally, all postcode level forecasts are aggregated to derive the regional solar forecast.

% as we use TCN as a building block in the proposed architectures, to evaluate the impact of reducing the number of TCN models required to forecast each individual power generation time series, the traditional TCN network is adapted for regional forecasting using a similar sub-region based approach we propose in \ref{sec:subregion_agg} for comparison. In this approach, a TCN is implemented for each sub-region in SWIS and trained with all the postcode power generation time series and weather data belonging to the sub region to forecast the power generation of each postcode in the sub region.

The following notations are used to describe the above-mentioned baseline approaches: 
\begin{enumerate}
    \item SN.Direct, SARIMA.Direct, LSTM.Direct, CNN.Direct, TCN.Direct refers to the baseline forecasting models implemented using regional forecasting strategy (i), 
    \item SN.PostcodeAGG, SARIMA.PostcodeAGG, LSTM.PostcodeAGG, CNN.PostcodeAGG, TCN.PostcodeAGG refers to baseline forecasting models implemented using regional forecasting strategy (ii), and
    \item TCN.Global.PostcodeAGG refers to the TCN implemented as a global forecasting model, and the postcode level forecasts are aggregated similar to the aggregation in strategy (ii)
\end{enumerate}

\subsection{Evaluation}
We evaluate the performance of all approaches described in this work using the last 36 days for testing and the rest of the data for training. All time series data is scaled with its mean and standard deviation of the training data before training the forecasting models. 

For deep learning methods, a part of the training data is used for validation and hyper-parameter tuning. There are several hyper-parameters we have tuned for the baseline forecasting models and our proposed architectures. A detailed description of these hyper-parameters is provided in \ref{appendixC}. The hyper-parameters for the deep learning models were tuned using a grid search algorithm \cite{Liashchynskyi2019GridNAS} in which we defined a candidate set of values for hyper-parameters and the best parameter combination is exhaustively searched. The models are trained to reduce the Mean Squared Error (MSE) loss with the Adam optimizer. To minimise over-fitting, an early stopping criteria is included in which the model training is stopped if the validation loss does not reduce over 50 consecutive epochs. Parameter tuning for SARIMA is done using the commonly used auto-arima approach. SARIMA was implemented using the \textit{pmdarima} package in Python and deep learning models including the proposed architectures were implemented with Tensorflow 2.0 and Python.

To evaluate the regional forecast performance the normalised root mean squared error (nRMSE) is used, which is a widely used evaluation metric in solar forecasting studies \cite{Yang2019OperationalMarket, Yagli2019AutomaticModels}. The nRMSE is calculated for each day, i.e, that is for each test sample in the testing period. The average nRMSE across the entire test period is then calculated to determine the final forecast errors. The nRMSE for a day-ahead forecast is calculated as shown in Equation \ref{eq:nrmse}, where $n$ is the number of time points of a day (18 time points in this case as night hours are excluded), $y_i$, $f_i$ are the actual regional power generation and forecast at time point $i$.

\begin{equation}
    nRMSE = \frac{\sqrt{\frac{1}{n}\sum_{i=1}^n(f_i - y_i)^2}}{\frac{1}{n}\sum_{i=1}^{n}y_i}
\label{eq:nrmse}
\end{equation}

Using the nRMSE the Forecast Skill Score  (SS) is further calculated to measure the quality of forecasts compared to a periodic persistence-based (i.e., SN) forecast \cite{Yang2019OperationalMarket}. The SS is calculated as shown in Equation \ref{eq:ss}.

\begin{equation}
    Skill Score (SS) = \left ( 1 - \frac{nRMSE_{forecasting \; method}}{nRMSE_{SN}}\right ) * 100\%
\label{eq:ss}
\end{equation}

To account for the randomness of deep learning models caused by random weight initialisation, all models are trained 10 times using 10 different seeds. The average nRMSE across the 10 runs is reported along with the standard deviation of the errors across the runs, where a lower standard deviation reflects a stable and consistent performance.

The Mann-Whitney U Test is used to evaluate the statistical significance of the forecast errors, which is a statistical test that determines whether two independent samples are significantly different from one another. We use a p-value$=0.05$, where a value less than 0.05 indicates the forecast errors produced by two models are significantly different from each other, whereas a value greater than 0.05 indicate otherwise.

\section{Results and Discussion}
\label{sec:results}

Table \ref{tab:table_results} shows the average nRMSE and Skill Score (\%) for day ahead forecasts generated by baseline and proposed forecasting models and regional forecasting strategies. The table summarises the type of input time series data used to train the forecasting model, the type of forecasting models, the forecasting strategy used to derive a regional forecast (as discussed in Section \ref{sec:baselines} and \ref{sec:direct_regional}, \ref{sec:subregion_agg}), the required number of models to generate a regional forecast for SWIS and the nRMSE and SS (\%) of the regional forecast. The training time associated with these approaches are shown in Table \ref{tab:run-times}.

\begin{table*}[!h]
\footnotesize
\centering
\begin{tabular}{clcrll} 
\hline
\begin{tabular}[c]{@{}c@{}}\textbf{Input}\\\textbf{Time Series}\end{tabular}                                               & \multicolumn{1}{c}{\begin{tabular}[c]{@{}c@{}}\textbf{Forecasting }\\\textbf{ Model}\end{tabular}} & \begin{tabular}[c]{@{}c@{}}\textbf{Forecasting }\\\textbf{ Strategy}\end{tabular}                                                              & \multicolumn{1}{c}{\begin{tabular}[c]{@{}c@{}}\textbf{Number }\textbf{ of }\\\textbf{ Models}\end{tabular}} & \multicolumn{1}{c}{\textbf{nRMSE}}  & \begin{tabular}[l]{@{}l@{}}\textbf{SS}\\\textbf{ (\%)}\end{tabular} \\ 
\hline
\multirow{5}{*}{\begin{tabular}[c]{@{}c@{}}Aggregated \\ Power Generation \\ Time Series\end{tabular}}                         & SN                                                                                                  & \multirow{5}{*}{\begin{tabular}[c]{@{}c@{}}Direct \\ Forecast\\ (Direct)\end{tabular}}  & 1 & 0.288* & - \\
    & SARIMA  & & 1 & 0.329* & -14.24 \\
    & LSTM  &  & 1  & 0.308 (± 0.005)* & -6.94 \\
    & CNN & & 1 & 0.305 (± 0.006)* & -5.90 \\
    & TCN & & 1  & 0.298 (± 0.015)* & -3.47 \\ 
\hline
\multirow{6}{*}{\begin{tabular}[c]{@{}c@{}} Postcode \\ Power Generation, \\ Weather \\ Time series\end{tabular}} & SN & \multirow{6}{*}{\begin{tabular}[c]{@{}c@{}}Aggregate \\ Postcode \\ Forecasts\\ (PostcodeAGG)\end{tabular}} & 101  & 0.288* & - \\
    & SARIMAX  & & 101 & 0.367* & -27.43 \\
    & LSTM  & & 101 & 0.218 (± 0.009)* & 24.31 \\
    & CNN & & 101  & 0.206 (± 0.004)* & 28.47 \\
    & TCN & & 101 & \textbf{0.184 (± 0.006)}  & 36.11 \\
    & TCN.Global & & 20 & 0.186 (±0.008) & 35.41 \\
\hline
\multirow{5}{*}{\begin{tabular}[c]{@{}c@{}}Aggregated, \\ Postcode Power\\ Generation and \\ Weather \\ Time Series\end{tabular}}  &  & \multirow{4}{*}{\begin{tabular}[c]{@{}l@{}}Direct \\ Forecast \\ (Direct)\end{tabular}} &   &  \\
    & HTCNN A1 & & 1 & 0.188 (± 0.015) & 34.72 \\
    & \\
    &   HTCNN A2 & & 1  & 0.187 (± 0.024)  & 35.06 \\
    &\\
\hline
\multirow{5}{*}{\begin{tabular}[c]{@{}c@{}}SubRegion, \\  Postcode Power \\ Generation \\ and Weather \\ Time Series\end{tabular}}  &  & \multirow{4}{*}{\begin{tabular}[c]{@{}c@{}}Aggregate\\ Subregion\\Forecasts\\ (SubRegionAGG)\end{tabular}} &   &  \\
    & HTCNN A1 & & 20 & 0.177 (± 0.008)   & 38.54 \\
    & \\
    & HTCNN A2 & & 20  & \textbf{0.172 (± 0.006)*} & 40.28  \\
    &\\
\hline
\end{tabular}
\caption{The nRMSE and Skill Score (SS) across the testing period for a day ahead forecast by baseline and proposed methods. For deep learning methods, the standard deviation across 10 different runs are shown next to the nRMSE value. Errors that are statistically significant from the best baseline (TCN.PostcodeAGG) are represented with an additional * sign next to the nRMSE.}
\label{tab:table_results}
\end{table*}

Table \ref{tab:table_results} clearly shows that when the aggregated power generation time series ($ts_{AGG}$) is used as the sole input for the forecasting models, deep learning models such as LSTM, CNN, and TCN do not yield significant improvements in forecasts compared to SN and SARIMA. Notably, SARIMA and SARIMAX perform poorly compared to all other models, including SN. Identifying the parameters for SARIMA/SARIMAX models manually can be challenging, especially when dealing with a large number of time series. To address this, we employ an automatic parameter identification technique called auto-arima in this study. The subpar performance of SARIMA models may be attributed to the automatic parameter identification process failing to capture the variability of the power generation time series data. Previous work has also reported similar observations, where ARIMA models underperformed compared to SN \cite{Trapero2015Short-termRegression}.

When the input consists of power generation data from postcodes along with associated weather data, regional-level forecasts exhibit improvements across all models, except for SN and SARIMAX. This observation highlights the capability of deep learning models to effectively capture the solar generation variability that is caused by weather patterns of each specific postcode. However, since SN does not utilize weather data, its approach of individually forecasting the generation of each postcode and then aggregating the results is equivalent to directly forecasting the aggregated time series. Therefore, the nRMSE for SN remains unchanged across the two approaches. On the other hand, SARIMAX incorporates weather data, yet its nRMSE has increased compared to other methods. This discrepancy might be attributed to the fact that SARIMAX entails 101 individual models, each requiring the identification of parameters for both the time series and the associated exogenous features. This process of automatic parameter identification becomes more challenging when dealing with a large number of individual time series, potentially leading to less accurate forecasts.

The best regional forecasts for SWIS from baseline approaches are achieved by forecasting each postcode power generation time series with TCN and aggregating the individual forecasts (TCN.PostcodeAGG). Moreover, convolution-based networks (CNN and TCN) have demonstrated superior performance compared to LSTM networks. This finding aligns with recent studies on day-ahead solar PV power forecasting, where convolution-based networks have been reported to outperform LSTMs  \cite{Lin2020TemporalForecasting, Wang2019ANetwork}. The strong performance of convolution-based networks can be attributed to their ability to capture relevant features from past power generation and weather data, which are crucial for predicting future power generation. When comparing the two convolution-based networks, CNN and TCN, it is evident that TCN exhibits significantly better performance than CNN. This observation confirms the effectiveness of temporal convolutions in forecasting time series data by capturing important temporal patterns, surpassing the capabilities of traditional 1D CNNs, as discussed in Section \ref{sec:method}.

When comparing the nRMSE of TCN and TCN.Global, it can be seen that TCN.Global exhibits a slightly higher nRMSE, with a marginal increase of 1\%, along with a larger standard deviation compared to TCN. However, there is no statistical significance observed between the two error distributions. It is important to note that TCN.Global has effectively reduced the number of models needed to forecast individual power generation time series data. This reduction in model count demonstrates the capability to achieve accurate regional forecasts while significantly reducing the required number of required models with a negligible loss in forecast accuracy.

\begin{figure}[!h]
    \centering
    \begin{subfigure}{\textwidth}
        \centering
         \includegraphics[width=\textwidth]{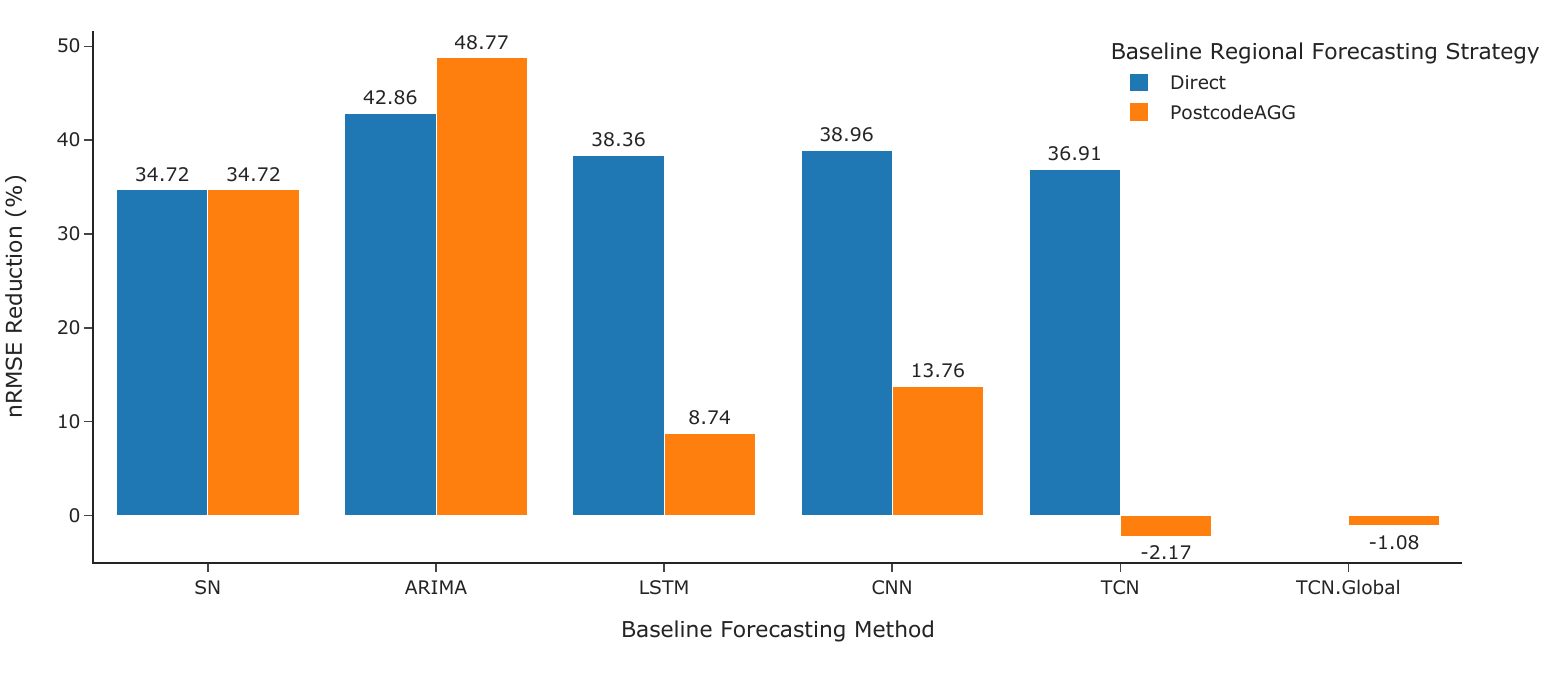}
         \caption{nRMSE reduction percentage using HTCNN A1-based direct strategy compared to baseline approaches.}
    \end{subfigure}
    \begin{subfigure}{\textwidth}
        \centering
         \includegraphics[width=\textwidth]{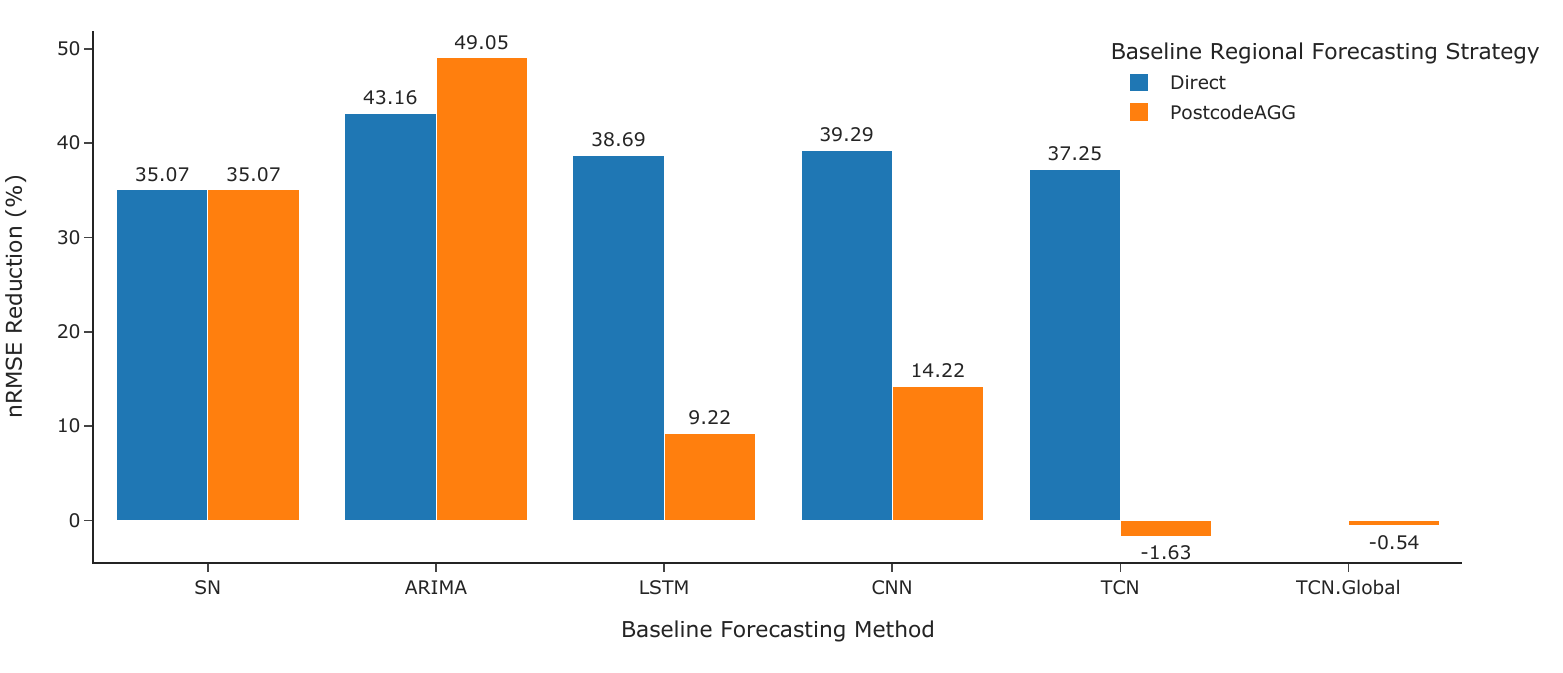}
         \caption{nRMSE reduction percentage using HTCNN A2-based direct strategy compared to baseline approaches.}
    \end{subfigure}
    \caption{nRMSE reduction percentages (\%) using \textbf{HTCNN-based methods} and \textbf{proposed direct forecasting strategy}.}
    \label{fig:direct_HTCNN}
\end{figure}

\begin{figure}[!h]
    \centering
    \begin{subfigure}{\textwidth}
        \centering
         \includegraphics[width=\textwidth]{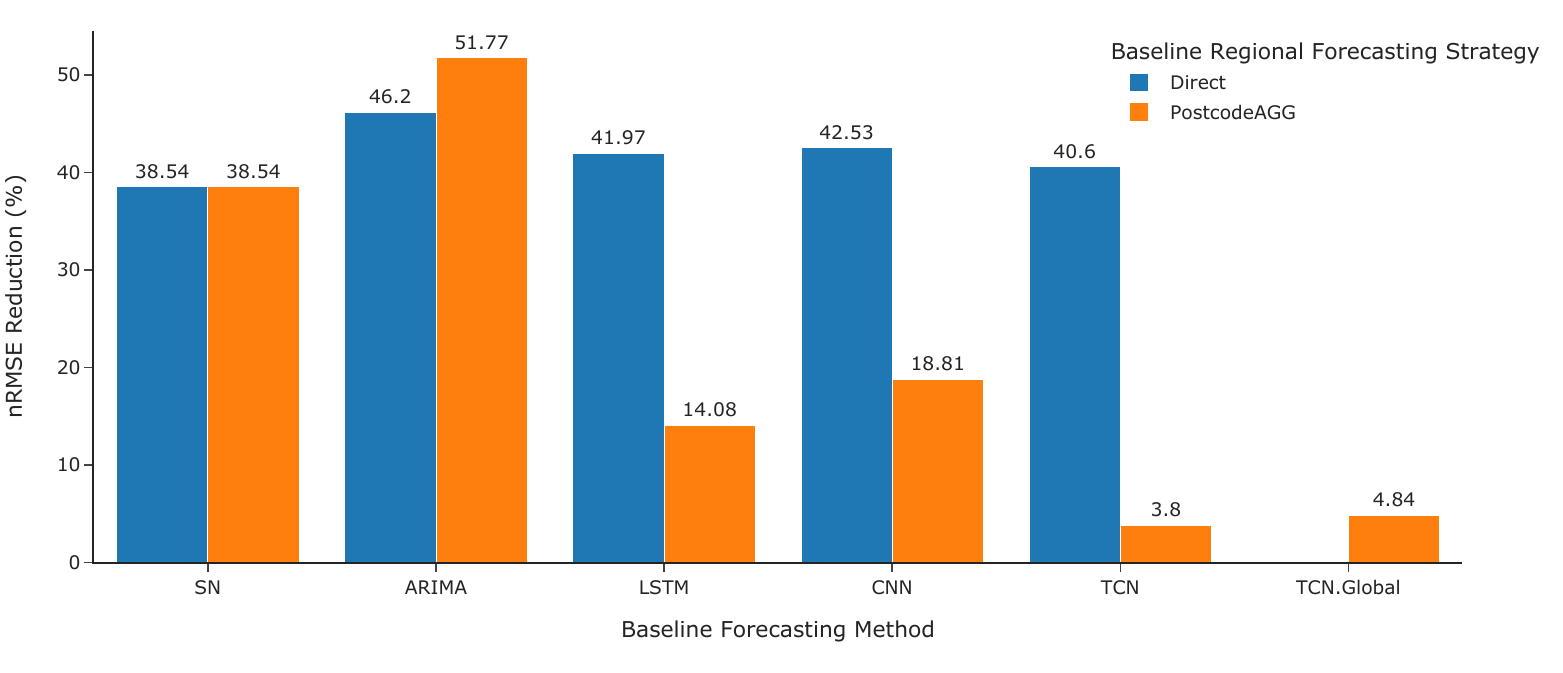}
         \caption{nRMSE reduction percentage using HTCNN A1-based sub-region forecast aggregation strategy compared to baseline approaches.}
    \end{subfigure}
    \begin{subfigure}{\textwidth}
        \centering
         \includegraphics[width=\textwidth]{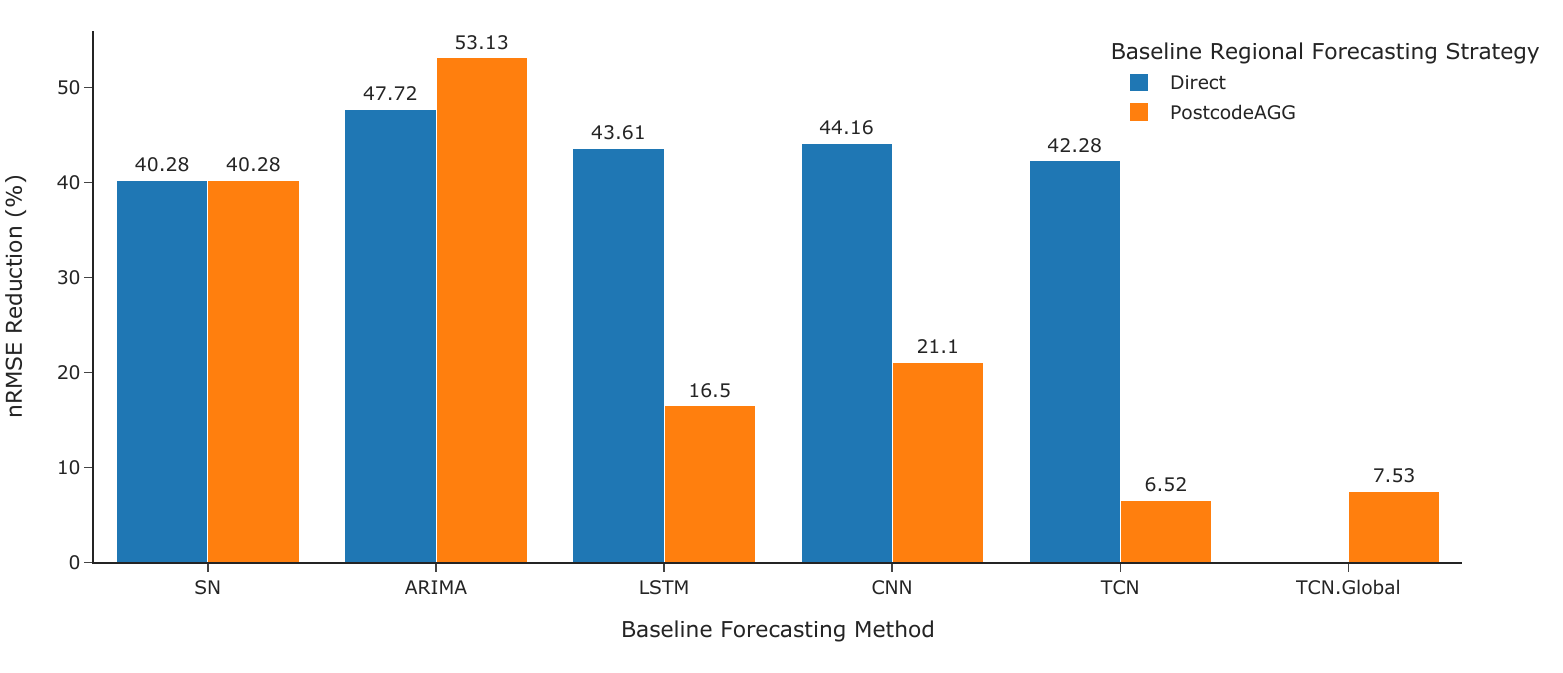}
         \caption{nRMSE reduction percentage using HTCNN A2-based sub-region forecast aggregation strategy compared to baseline approaches.}
    \end{subfigure}
    \caption{nRMSE reduction percentages (\%) using \textbf{HTCNN-based methods} and \textbf{proposed sub-region forecast aggregation strategy}.}
    \label{fig:Subregion-HTCNN}
\end{figure}

Figures \ref{fig:direct_HTCNN}, and \ref{fig:Subregion-HTCNN} illustrate the nRMSE reduction percentages achieved by the proposed approaches compared to the baseline approaches \footnote{nRMSE reduction percentage calculated as $\frac{nRMSE_{baseline} - nRMSE_{proposed}}{nRMSE_{baseline}} * 100\%$}. Figure \ref{fig:direct_HTCNN} shows that when directly forecasting the SWIS generation using HTCNN A1 or HTCNN A2, the nRMSE is reduced by 8.7\% to 49\% compared to most baselines, with the exception of TCN.PostcodeAGG and TCN.Global.PostcodeAGG. HTCNN A1 and HTCNN A2 exhibit a slight increase in forecast error, with 2.17\% and 1.63\% respectively, when compared to the best-performing baseline, TCN.PostcodeAGG. However, these errors are not statistically significant. The better performance of TCN.PostcodeAGG can be attributed to the smoothing effect discussed in Section \ref{sec:related}, which occurs when combining individual forecasts. In this approach, any overestimation for one time series is compensated by the underestimation for another during the aggregation process. However, in comparison to TCN.PostcodeAGG, which requires 101 models, HTCNN A1 and HTCNN A2, employing the direct forecast strategy, require only one network to achieve a similar level of forecast accuracy. Modelling hundreds of diverse time series originating from geographically dispersed locations in a region can still be difficult for a single HTCNN, regardless of its architecture and therefore contributing to the marginal increase in forecast error observed in this case.

%  The superior performance of TCN.PostcodeAGG may be due to the smoothing effect (discussed in Section \ref{sec:related}) caused when adding individual forecasts, as an overestimation of a forecast for one time series will be compensated by the underestimation for another during addition. However, it is noteworthy that TCN.PostcodeAGG has 101 individually trained networks specialised to forecast the power generation of each postcode separately, while HTCNN A1 and HTCNN A2 with the direct forecast strategy only requires one network to achieve a similar forecast accuracy. Modelling hundreds of diverse time series coming from geographically dispersed locations in a region can still be challenging with a single HTCNN regardless of the architecture and therefore, leading to the marginal increase in the forecast error.

As shown in Figure \ref{fig:Subregion-HTCNN}, the sub-region forecast aggregation strategy (SubRegionAGG) employed by HTCNN A1 and A2 results in a reduction of nRMSE between 3.8\% and 53.13\% compared to all baselines. Notably, the nRMSE is reduced by 3.8\% and 6.5\% when compared to the best performing baseline, TCN.PostcodeAGG. While the 3.8\% reduction achieved by HTCNN A1 is not statistically significant, the 6.5\% reduction achieved by HTCNN A2 is statistically significant. It can be further seen that HTCNN.A2.SubRegionAGG achieves the highest SS of 40.2\%. This indicates that HTCNN.A2.SubRegionAGG outperforms other models in forecasting the regional solar generation of SWIS.

The superior performance of HTCNN A2 may be attributed to two factors. Firstly, compared to HTCNN A1, HTCNN A2 effectively utilizes features captured from the individual time series throughout the network. These features encompass the influence of weather factors, and reusing such essential features within the network allows for better incorporation of these factors in the forecasting task. Secondly, HTCNN.A2.SubRegionAGG is trained with time series data from nearby locations that are affected by similar weather conditions. This is advantageous as the network only needs to learn from data that exhibits similar generation profiles, thereby simplifying the learning task. This observation aligns with the findings presented in Table \ref{tab:table_results}, where both HTCNN.A1.SubRegionAGG and HTCNN.A2.SubRegionAGG significantly reduce the nRMSE compared to HTCNNA1.Direct and HTCNNA2.Direct.

\begin{figure}[]
     \centering
     \begin{subfigure}[]{\textwidth}
         \centering
         \includegraphics[width=0.95\textwidth]{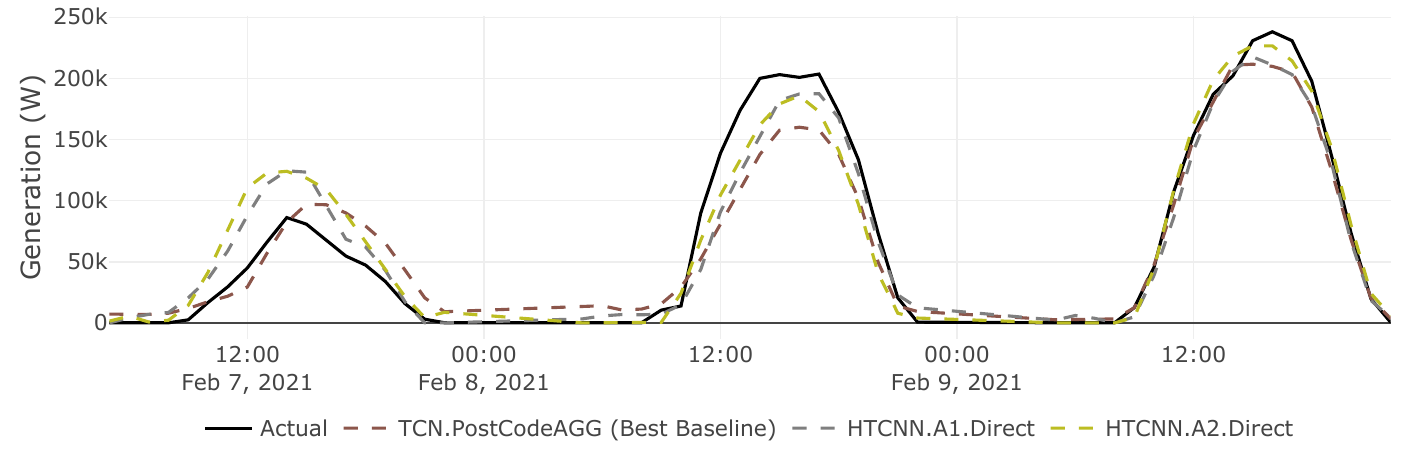}
         \caption{HTCNNs with the direct forecasting strategy.}
         \label{fig:fc_plot}
     \end{subfigure}
     \hfill
     \begin{subfigure}[]{\textwidth}
         \centering
         \includegraphics[width=0.95\textwidth]{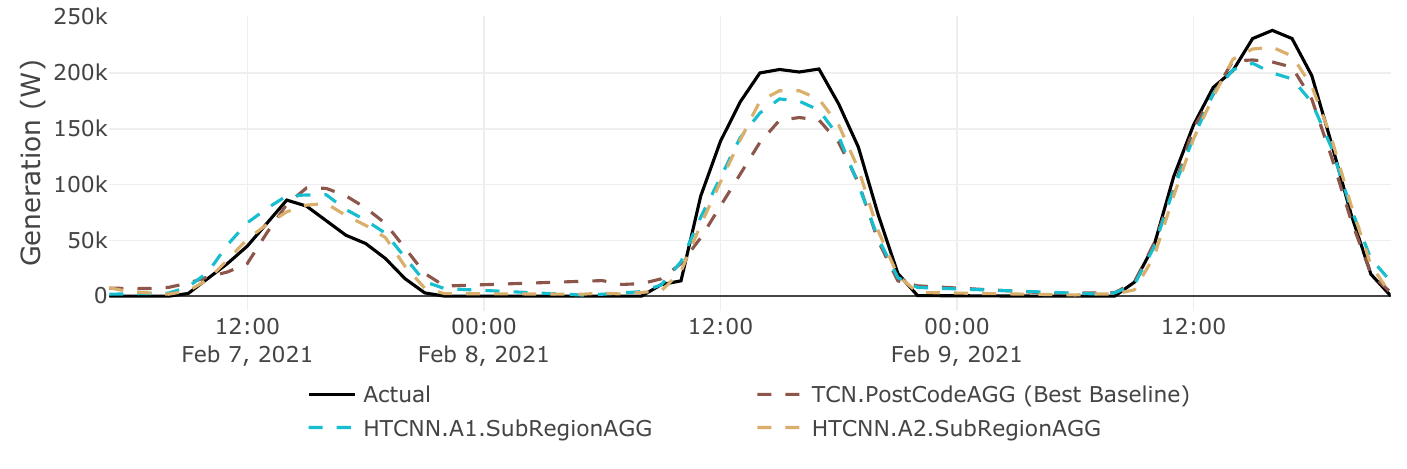}
         \caption{HTCNNs with the sub-region forecast aggregation strategy.}
         \label{fig:fc_plot2}
     \end{subfigure}
    \caption{Day ahead forecasts for three test days by the best-performing baseline approach (TCN.PostCodeAGG) and proposed approaches.}
    \label{fig:fc_plots}
\end{figure}

Observing the forecasts from the best performing baseline - TCN.PostCodeAGG and the proposed approaches as shown in Figure \ref{fig:fc_plots} for three test days, it can be seen that the  HTCNN.A1.SubRegionAGG, and HTCNN.A2.SubRegionAGG provide forecasts closer to the actual generation. HTCNN.A1.Direct and HTCNN.A2.Direct perform similarly except in some cases such as the solar generation profile on February 7. However, even on this cloudy day (February 7) HTCNN.A1.Direct and HTCNN.A2.Direct are able to identify the low solar generation profile of the region. 

% \begin{figure}
%     \centering
%     \includegraphics[width=\textwidth]{figures/fc_plot2.pdf}
%     \caption{Day ahead forecasts for three test days by the best performing baseline - TCN.PostCodeAGG and proposed approaches.}
%     \label{fig:fc_plots}
% \end{figure}

\begin{figure*}[]
    \centering
    \includegraphics[width=\textwidth]{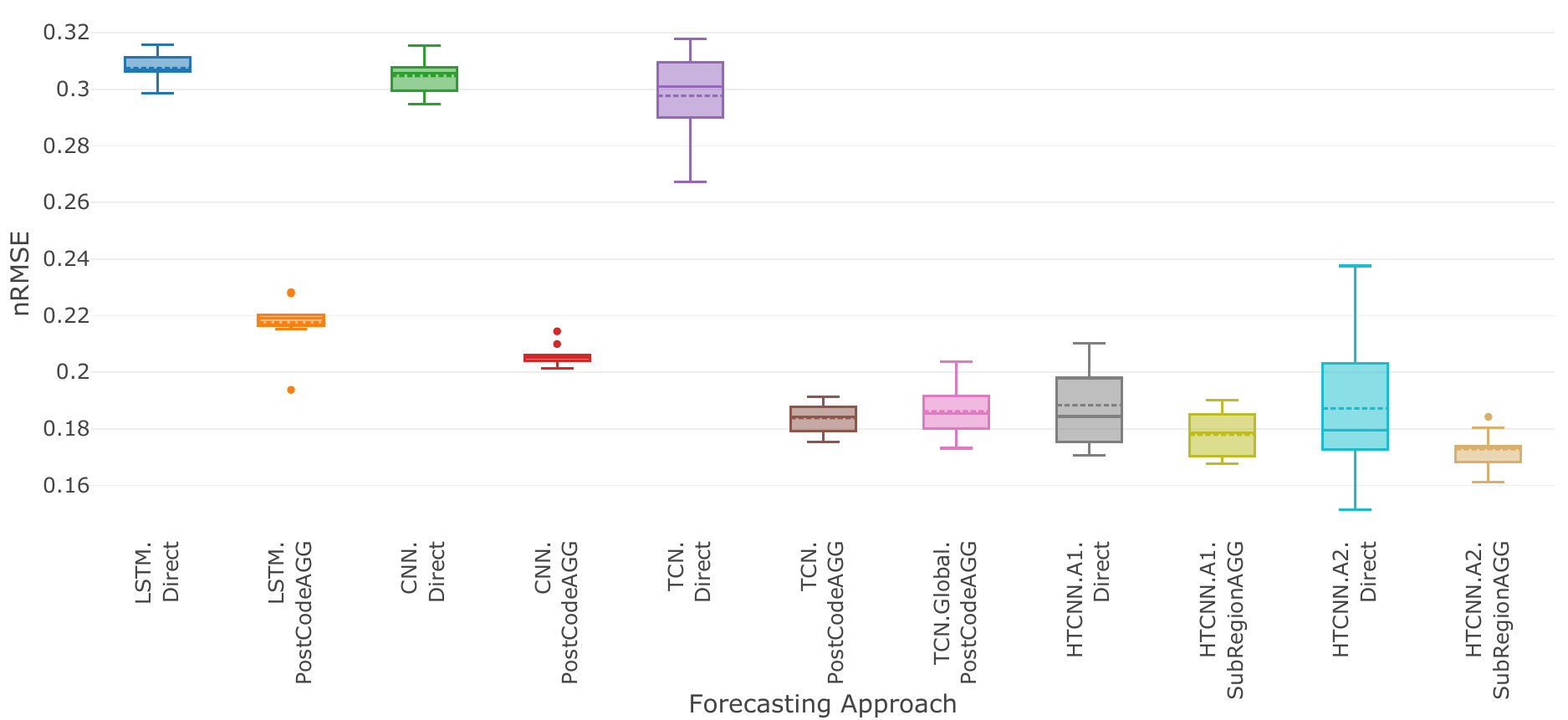}
    \caption{nRMSE distribution of the test samples in 10 different runs for deep learning-based approaches.}
    \label{fig:distplot}
\end{figure*}

Figure \ref{fig:distplot} presents the distribution of nRMSE values for 10 different runs of the deep learning-based approaches. It can be seen that TCN.Direct, HTCNN.A1.Direct, and HTCNN.A2.Direct exhibit higher variability in nRMSE values across the 10 runs compared to other approaches. This indicates that these models are more prone to generating inconsistent forecasts due to the weights initialization of the networks. In contrast, the best performing baseline, TCN.PostcodeAGG, and the proposed approaches HTCNN.A1.SubRegionAGG and HTCNN.A2.SubRegionAGG have a consistent forecast performance regardless of the initial weights of the networks. These findings highlight that the overall best performing approach for regional forecasting proposed in this work, namely HTCNN.A2.SubRegionAGG, not only provides accurate forecasts but also ensures a consistent forecast performance.

\textit{The value of solar forecast improvements:} The proposed HTCNN.A2.SubRegionAGG brings valuable improvements in solar forecasting, with a 40.2\% increase in the Skill Score and a 6.5\% reduction in nRMSE compared to the best baseline. Beyond these statistical measures, such improvements translate into tangible economic benefits for stakeholders in the energy sector \cite{wang2022cost, gandhi2024value}. Accurate PV power forecasts enable grid operators to manage electricity supply and demand effectively and ensure the allocation of reserves to maintain power system reliability \cite{Antonanzas2016ReviewForecasting}. Even small errors in solar forecasts can lead to significant costs for power systems operations, impacting electricity load, price forecasting, and the flexibility of supply and demand. For example, recent research suggests with every 10\% increase in the forecast skill score, the value gained from solar forecasts can range from 0.07-3.31 USD/MWh depending on the specific application for which these forecasts are utilised \cite{gandhi2024value}. Applying this insight to a region like SWIS, which generates 18 million MWh of electricity annually, the 6.5\% error reduction can translate to a significant economic value in thousands of dollars. This emphasises the practical and economic implications of solar forecast improvements achieved from HTCNN.A2.SubRegionAGG, for optimal decision-making and operational efficiency in power systems.

\textit{Generalisability under different climatic conditions:} The proposed work analysed 101 PV sites located in the southern and western regions of Western Australia, which was chosen as a case study for regional solar forecasting.  In these areas, the annual average daily solar exposure, quantified as total solar energy (typically ranging from 1 to 35 $MJ/m^2$), fluctuates between 15-21 $MJ/m^2$, with peaks reaching up to 35 $MJ/m^2$ during the summer months \cite{bom}. The solar exposure, and consequently the PV power output, is influenced by cloud cover, with regions closer to the coast in southern WA experiencing lower solar exposure due to increased cloud cover. Coastal areas, particularly in the south, exhibit higher moisture content (higher relative humidity), leading to more extensive and frequent cloud cover and, consequently, reduced solar exposure. To account for these diverse weather conditions, the proposed work incorporates seven weather features: UV index, cloud cover, humidity, pressure, wind speed, temperature, and dew point. As weather variations in different regions are captured using such external variables, the proposed HTCNN-based approaches can be applied in regions with similar or different solar exposures through additional training with data specific to those regions. However, based on the results it was seen that when HTCNN was trained using the sub-region forecasting strategy, wherein the model was trained using data from PV sites influenced by similar weather conditions, it performed significantly better compared to the direct forecasting strategy. Therefore, for larger regions where different parts of the region may be affected by large weather variations, it is more suitable to consider a sub-region-based strategy for regional solar forecasting.

\subsection{Limitations and Future Improvements}

In this study, we have focused on achieving accurate regional forecasts by leveraging both aggregated and individual (postcode-level) time series within a region. Our goal was to improve the forecast accuracy while also minimising the number of required models. To fulfill this objective, we have conducted comparisons with deep learning-based methods that are widely explored and have undergone extensive validation in solar forecasting literature. However, more recently transformer architecture variants \cite{zhou2022fedformer} have shown success in time series forecasting tasks and have also been adopted in solar power forecasting studies \cite{liu2023transformer}. Recent literature in the time series forecasting domain has further introduced linear models \cite{zeng2023transformers, chen2023tsmixer} questioning the effectiveness of transformer-based methods for time series forecasting tasks. In this work, we have not examined these recent methods as we focused on establishing a solid baseline and understanding the performance characteristics of well-established models in solar forecasting literature. However, evaluating such alternative methods could potentially offer novel insights and perspectives, contributing to a more comprehensive understanding of the accuracies that can be achieved when forecasting solar generation at a regional level.

As discussed in Sections \ref{subsec:HTCNNA1} and \ref{subsec:HTCNNA2}, the proposed HTCNN networks incorporate distinct paths within the network to learn from both the aggregated and multiple individual (postcode-level) time series in a region. During network training, both types of time series were processed together, utilizing the entire dataset. However, when dealing with a large number of time series, such a training approach can be computationally expensive. Alternatively, these networks can be trained as separate sub-networks and later combined to create a similar network, potentially reducing the training overheads. Exploring training strategies, such as dividing the architecture into sub-networks, training them individually, and subsequently combining the pre-trained networks, would be an interesting avenue for future research. Furthermore, it would be valuable to investigate the impact of such alternative training strategies on the overall forecast performance and training times of the HTCNNs.

When new data becomes available due to the installation of new PV systems, the proposed HTCNN-based approaches would typically require retraining from scratch to account for this additional data. This process can introduce additional training costs. Therefore, investigating how HTCNNs can be adapted without retraining from scratch (e.g., continual or incremental learning of networks \cite{van2022three, melgar2023identifying}) would be an important direction for future work. 

This work primarily focused on rooftop PV systems, which have experienced significant growth in recent years reshaping the traditional form of power supply and demand. Although our investigation specifically targeted rooftop PV systems within a given region for the purpose of regional solar forecasting, it is important to note that any behind-the-meter PV generation source can influence the overall regional solar generation. For example, Building Integrated Photovoltaic Systems (BIPV) \cite{liu2023review} are an additional behind-the-meter power generation source that impacts regional solar generation. While the proposed work can be extended considering BIPV systems, exploring how the generation from BIPV systems affects regional forecasting is a potential avenue of future research.

\section{Conclusion}
\label{sec:conclusion}

With the increasing uptake of rooftop solar PV installations, regional solar power forecasts that take into account the solar generation from all PV systems within a region are becoming important for many stakeholders of the grid to ensure a stable electricity supply. In this work, deep learning-based regional solar power forecasting approaches are proposed. We introduced hierarchical temporal convolutional neural networks (HTCNNs) to forecast the aggregate solar power generation of an entire region. HTCNNs were trained using all power generation time series in a region, including weather data collected from multiple geo-spatially diverse locations in the region. Two networks HTCNN A1 and A2 which have different network structures, were proposed for this task. We further introduced and discussed two strategies on how HTCNNs can be used for regional solar power forecasting. The first strategy directly predicted the power generation of the entire region using a single HTCNN. The second divided the larger region into sub-regions and trained an individual HTCNN for each of these, and the forecasts were aggregated to derive a regional forecast.

The proposed approaches were validated using a large distributed solar PV power dataset collected from 101 locations in Western Australia and were evaluated against multiple baseline forecasting models: seasonal naive, ARIMA, LSTM, traditional 1D CNN, and TCN.  In comparison to the best-performing baseline method (TCN), HTConveNet A2 sub-region based method reduced a statistically significant normalised root mean squared error (nRMSE) of 6.5\%. Furthermore, in comparison to all baseline forecasting approaches, using the sub-region based forecasting strategy, any HTCNN network reduced the nRMSE by 3.8\% or more. Using the direct forecasting strategy, the nRMSE was reduced by 8.7\% or more compared to most baseline approaches except two that were based on TCNs. It was further demonstrated that all HTCNN-based approaches required fewer networks to forecast the regional solar generation compared to existing approaches.

% \section{Data Availability}

\section*{Declaration of Competing Interest}
The authors declare that they have no known competing financial interests or personal relationships that could have appeared to influence the work reported in this paper.

\section*{Acknowledgement}
The authors are grateful to Solar Analytics for providing anonymised solar power generation data to conduct this research. This work was supported by the Melbourne Research Scholarship awarded to the first author. SH and JDH acknowledge Australian Research Council grant DP220101035. The authors acknowledge the support of Tamasha Malepathirana, Rashindrie Perera, Ken Chen and Will Bodewes for their valuable feedback and input on the paper. The research was undertaken using the LIEF HPC-GPGPU Facility hosted at the University of Melbourne (this facility was established with the assistance of LIEF Grant LE170100200). 

\appendix
\section{Standard 1D Convolution vs Dilated 1D Convolution}
\label{appendixA}

Figure \ref{fig:convolutionOperation} shows the an example of the standard 1D convolution and dilated convolution. The convolution operation can be causal or non-causal. The following shows causal convolution where the output $y_t$ at time t is convolved only with input from time t and earlier. As can be seen, if the dilation rate is 1 this is the same as standard convolution.

\setcounter{figure}{0}    
\begin{figure}[!h]
     \centering
     \begin{subfigure}[b]{0.5\textwidth}
         \centering
         \includegraphics[width=\textwidth]{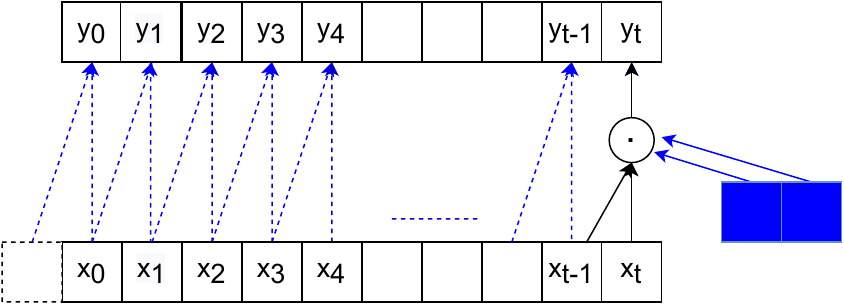}
         \caption{1D causal convolution operation. Blue square shows a filter of size 2. $\cdot$ shows the dot product. The blue dash arrows show the compact representation of convolving the filter across the input.}
         \label{fig:convExample}
     \end{subfigure}
     \hfill
     \begin{subfigure}[b]{0.5\textwidth}
         \centering
         \includegraphics[width=\textwidth]{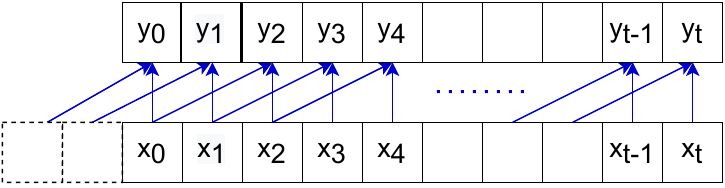}
         \caption{1D dilated (dilation rate = 2) causal convolution operation. The blue arrows show the compact representation of convolving the filter across the input.}
         \label{fig:dilatedconvExample}
     \end{subfigure}
    \caption{Standard 1D causal convolution and dilated 1D causal convolution with a dilation rate of 2 on an input sequence $x_0, x_1, x_2,\dots,x_{t-1},x_t$. The output feature map generated after applying the filter is shown in $y$.}
    \label{fig:convolutionOperation}
\end{figure}

\section{Long-Short Term Memory Networks}
\label{appendixB}

LSTM networks consists of four main components: input gate ($i_t$), output gate ($o_t)$, forget gate ($f_t)$, cell state ($C_t$) and the hidden state ($h_t$) with regard to a time point $t$. The input gate decides what new information is added, the output gate decides the next hidden state, the forget gate decides what information to be forgotten/ kept, and the cell state carries the relevant information throughout the processing of the input sequence. Equation \ref{eq:subeq1}-\ref{eq:subeq4} shows the operations of these components. Considering a hidden state with cell dimension $d$ and an input $x_t$, $W_{ih}, W_{oh}, W_{fh}, W_{ch}, W_{ix}, W_{ox},  W_{fx}, W_{cx}$ are the weight matrices of the input gate, forget gate, output gate and cell state associated with the hidden state and input. $b_{i}, b_{o}, b_{f}, b_{c}$ are bias vectors. $\cdot, \odot$ is the point-wise and element-wise multiplications. $\sigma$ and $\tanh$, denotes the sigmoid (squishes values between 0 and 1) and tanh (squishes values between -1 and 1) activation. Gates contains $\sigma$ activation, which is helpful to update or forget data because any number getting multiplied by 0 is 0, causing values to be forgotten. Any number multiplied by 1 is the same value therefore that value stays the same (i.e. preserved).

\begin{subequations}
\begin{align}
    i_t = \sigma(W_{ih} \cdot h_{t-1} + W_{ix} \cdot x_{t} + b_{i})  \label{eq:subeq1} \\
    o_t = \sigma(W_{oh} \cdot h_{t-1} + W_{ox} \cdot x_{t} + b_{o}) \label{eq:subeq2}\\
    f_t = \sigma(W_{fh} \cdot h_{t-1} + W_{fx} \cdot x_{t} + b_{f}) \label{eq:subeq3}\\
    C_t = i_t \odot \tanh(W_{ch} \cdot h_{t-1} + W_{cx} \cdot x_{t} + b_{c}) + f_{t} \cdot C_{t-1} \\
    h_{t} = o_{t} \odot \tanh(C_t) \label{eq:subeq4}
\end{align}
\label{eq:lstm}
\end{subequations}

\section{Hyper parameters}
\label{appendixC}

\begin{table*}[!ht]
\footnotesize
\centering
\begin{tabular}{ll} 
\hline

Forecasting Model & Hyper Parameters \\ 
\hline
SARIMA                                & \begin{tabular}[c]{@{}l@{}}$p$ - non seasonal AR term\\$q$ - non seasonal MA term\\$d$- non seasonal difference\\$P$ - seasonal AR term\\$Q$ - seasonal MA term\\$D$ - seasonal difference\end{tabular}                                                                                                                           \\ 
\hline
LSTM                                  & \begin{tabular}[c]{@{}l@{}} $d$ - Cell dimension \\ Number of stacked LSTM layers\\Number of epochs, Batch size, Learning rate\end{tabular}                                                                                                                                                                                           \\ 
\hline
CNN                                   & \begin{tabular}[c]{@{}l@{}}Filter size, Number of filters\\Number of 1D convolution and pooling layers\\Number of epochs, Batch size, Learning rate\end{tabular}                                                                                                                                                       \\ 
\hline
TCN                                   & \begin{tabular}[c]{@{}l@{}}Filter size, Number of filters\\$m$ - Number of Residual Blocks\\Number of epochs, Batch size\\Learning rate, Drop out rate\end{tabular}                                                                                                                                                      \\ 
\hline
HTCNN A1                          & \begin{tabular}[c]{@{}l@{}}Filter size\\$F'$, $F''$ - Number of filters in Convolution Stage (1), (2)\\Number of TCN blocks in Convolution Stage (1), (2), \\ Number of Residual blocks within a TCN block\\Number of epochs, Batch size, Learning rate\end{tabular}                                                                                                        \\ 
\hline
HTCNN A2                          & \begin{tabular}[c]{@{}l@{}}Filter size\\$F'$ - Number of Filters in Convolution Stage (1)\\Number of TCN blocks in Convolution Stage (1)\\ Number of Residual blocks within a TCN block\\$k$ - Number of Convolution and Concatenation layers\\$F''$ - Number of filters in~Convolution and Concatenation layers\\Number of epochs, Batch size, Learning rate\end{tabular}  \\
\hline
\end{tabular}
\caption{Hyper parameters that were tuned for different forecasting models discussed in this work.}
\label{tab:hyperparams}
\end{table*}

\newpage

\section{Training Times of Proposed and Baseline Approaches}
\label{appendixD}

Table \ref{tab:run-times} shows the training time for the proposed and baseline deep learning-based approaches. It can be seen that TCN-based approaches require more training time compared to LSTM and CNN-based approaches. Furthermore, although the proposed HTCNN.A1.Direct, HTCNN.A2.Direct, HTCNN.A1.SubRegionAGG, HTCNN.A2.SubRegionAGG approaches have higher training times per model, these approaches have significantly reduced the number of models that are required for regional solar forecasting and thereby, reducing the total time required to achieve an accurate regional forecast.

\begin{table}[!h]
\footnotesize
    \centering
    \begin{tabular}{p{5cm}p{4cm}p{2cm}p{2cm}}
    \hline
         Forecasting Approach & Training Time (H:M) per Model & Number of Models \\
         \hline
         LSTM.Direct & 00:25 & 1 \\
         CNN.Direct & 00:19 & 1 \\
         TCN.Direct & 01:07 & 1 \\
         LSTM.PostcodeAGG & 01:39 & 101 \\ 
         CNN.PostcodeAGG & 01:44 & 101 \\
         TCN.PostcodeAGG & 02:08 & 101 \\
         TCN.Global & 02:35 & 20 \\
         HTCNN.A1.Direct & 03:47 & 1 \\
         HTCNN.A2.Direct & 05:19 & 1\\
         HTCNN.A1.SubRegionAGG & 03:25 & 20 \\
         HTCNN.A2.SubRegionAGG & 04:16 & 20\\
         \hline
    \end{tabular}
    \caption{Training time for proposed and baseline approaches. Training time per model is presented as H (Hours) and M (Minutes) on a Nvidia A100 GPU.}
    \label{tab:run-times}
\end{table}



% \section*{CRediT authorship contribution statement}

\bibliographystyle{elsarticle-num-names} 
\bibliography{ref-external.bib, main_references.bib}





% \end{thebibliography}

\end{document}